\documentclass[final,5p,times,twocolumn,authoryear]{elsarticle}

\usepackage{amssymb}
\usepackage{algorithm}
\usepackage{algpseudocode}
\usepackage{graphicx}

\usepackage{subfigure}
\usepackage{amsmath}

\usepackage{amssymb}
\usepackage{xcolor}
\usepackage{tabularx} 
\usepackage{multirow}
\usepackage{tabu}
\usepackage{pbox}
\usepackage{amssymb}
\usepackage{amsfonts}
\usepackage{bbm}
\usepackage{graphicx}
{\begin{list}               
    {$\bullet$ \hfill}{
        \setlength{\leftmargin}{\parindent}
        \setlength{\parsep}{0.04\baselineskip}
        \setlength{\itemsep}{0.5\parsep}
        \setlength{\labelwidth}{\leftmargin}
        \setlength{\labelsep}{0em}}
    }
{\end{list}}

\providecommand{\eref}[1]{Eq. \eqref{#1}}  
\providecommand{\cref}[1]{Chapter~\ref{#1}}
\providecommand{\sref}[1]{Section~\ref{#1}}
\providecommand{\fref}[1]{Figure~\ref{#1}}
\providecommand{\tref}[1]{Table~\ref{#1}}

\providecommand{\norm}[1]{\lVert#1\rVert}

\renewcommand{\vec}[1]{\ensuremath{\boldsymbol{#1}}}
\providecommand{\mat}[1]{\ensuremath{\boldsymbol{#1}}}

\providecommand{\calL}{\mathcal{L}}

\providecommand{\mA}{\mat{A}}
\providecommand{\mB}{\mat{B}}
\providecommand{\mC}{\mat{C}}

\providecommand{\mH}{\mat{H}}
\providecommand{\mI}{\mat{I}}

\providecommand{\mP}{\mat{P}}

\providecommand{\mS}{\mat{S}}

\providecommand{\mW}{\mat{W}}

\providecommand{\ve}{\vec{e}}

\providecommand{\vv}{\vec{v}}

\newcommand{\argmax}[1]{\mathop{\underset{#1}{\mbox{argmax}}}}

\journal{Engineering Applications of Artificial Intelligence}

\begin{document}

\begin{frontmatter}

\title{Pixel-Level Clustering Network for Unsupervised Image Segmentation}

\author{Cuong Manh Hoang}
\ead{cuonghoang@seoultech.ac.kr}

\author{Byeongkeun Kang\corref{cor1}}
\ead{byeongkeun.kang@seoultech.ac.kr}
\cortext[cor1]{Corresponding author.}
\affiliation[ ]{organization={Department of Electronic Engineering, Seoul National University of Science and Technology},
            addressline={232 Gongneung-ro, Nowon-gu}, 
            city={Seoul},
            postcode={01811}, 
            country={South Korea}}

\begin{abstract}
While image segmentation is crucial in various computer vision applications, such as autonomous driving, grasping, and robot navigation, annotating all objects at the pixel-level for training is nearly impossible. Therefore, the study of unsupervised image segmentation methods is essential. In this paper, we present a pixel-level clustering framework for segmenting images into regions without using ground truth annotations. The proposed framework includes feature embedding modules with an attention mechanism, a feature statistics computing module, image reconstruction, and superpixel segmentation to achieve accurate unsupervised segmentation. Additionally, we propose a training strategy that utilizes intra-consistency within each superpixel, inter-similarity/dissimilarity between neighboring superpixels, and structural similarity between images. To avoid potential over-segmentation caused by superpixel-based losses, we also propose a post-processing method. Furthermore, we present an extension of the proposed method for unsupervised semantic segmentation. We conducted experiments on three publicly available datasets (Berkeley segmentation dataset, PASCAL VOC 2012 dataset, and COCO-Stuff dataset) to demonstrate the effectiveness of the proposed framework. The experimental results show that the proposed framework outperforms previous state-of-the-art methods.
\end{abstract}


\end{frontmatter}



\section{INTRODUCTION}
Image segmentation is an essential task in various computer vision applications, where it groups the pixels of an input image into segments that contain pixels belonging to the same object, stuff, or component. This task plays a critical role in numerous fields, including robot navigation, grasping, and autonomous driving. 
Specifically, in robot navigation and autonomous driving, image segmentation is used to distinguish between ground, objects, and air in an image, ensuring safe navigation. In robot grasping, image segmentation helps in localizing the target object among other objects.

While supervised semantic segmentation has attracted significant attention from researchers because of its high accuracy \citep{shelhamer2017, kang2019, Unet2023EFASPP}, these methods typically require a large dataset consisting of images and their pixel-level class labels, which is expensive to obtain. Moreover, supervised semantic segmentation methods have limitations in handling unknown object classes since they are constrained to classify each pixel into one of the predefined categories. For instance, when a robot encounters an object that does not belong to any of the predefined classes, it is impossible to segment the object correctly.

To avoid costly pixel-level annotations and limitations of predefined classes, researchers have explored unsupervised image segmentation and unsupervised semantic segmentation~\citep{IIC2019, clustering2022RGBD}. While unsupervised semantic segmentation usually requires a set of training images, unsupervised image segmentation can operate with only one image. This is because unsupervised semantic segmentation aims to group pixels of the same object or stuff type across multiple images. In this paper, we investigate an unsupervised image segmentation method that can segment any object or stuff using only one image and without the limitations of predefined classes.

Various pixel-level prediction tasks have achieved high accuracy through CNN-based methods~\citep{kang2018, Nakajima2019}. Consequently, unsupervised image segmentation methods have also incorporated CNNs~\citep{xia2017w, Kanezaki2018}. Since these methods lack access to annotations, \cite{xia2017w} proposed using image reconstruction to guide feature embedding for an image. Alternatively, \cite{Kanezaki2018} proposed using superpixels to compute a loss. Since then, various approaches have been proposed to improve unsupervised image segmentation by avoiding using superpixel segmentations \citep{kim2020unsupervised, kim2020}, utilizing both image reconstruction and superpixel segmentation \citep{Lin2020}, and by training an additional clustering module~\citep{dic2020}. However, they often suffer from over-segmentation due to inconsistent features within an object caused by the losses using either superpixels or only neighboring pixels.

\begin{figure}[!t] \begin{center}
\begin{minipage}{0.49\linewidth}
\centerline{\includegraphics[scale=0.62]{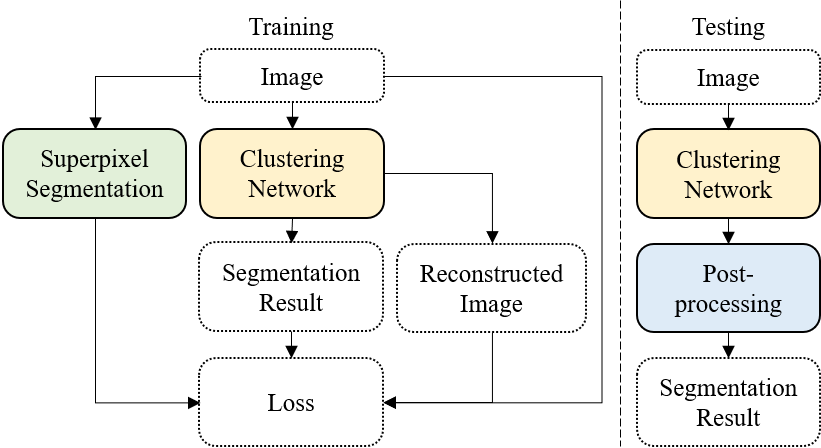}}
\end{minipage}
   \caption{Illustration of the proposed framework. In training, an input image is fed into both a clustering network and a superpixel segmentation algorithm. Then, a loss is computed by using the segmentation result, superpixels, reconstructed image, and input image. In testing, superpixel segmentation and image reconstruction are not used. Hence, an input image is fed into only the clustering network to obtain an initial segmentation result. It is further processed by a post-processing method to obtain the final segmentation result.}
\label{fig:overview}
\end{center}\end{figure}

\begin{figure}[!t] \begin{center}
\begin{minipage}{0.46\linewidth}
\centerline{\includegraphics[scale=0.22]{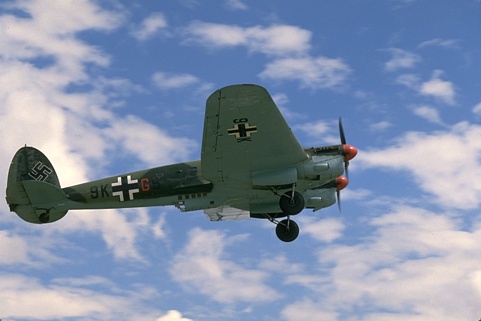}}
\end{minipage}
\begin{minipage}{0.46\linewidth}
\centerline{\includegraphics[scale=0.22]{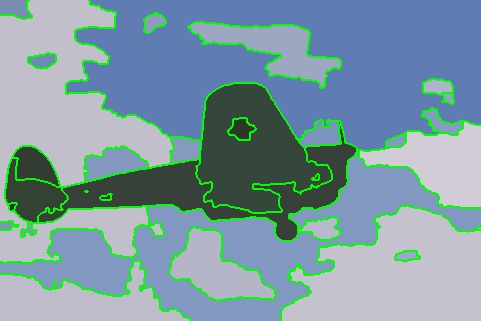}}
\end{minipage}
\\
\begin{minipage}{0.46\linewidth}
\centerline{(a)}
\end{minipage}
\begin{minipage}{0.46\linewidth}
\centerline{(b)}
\end{minipage}
\\
\begin{minipage}{0.46\linewidth}
\centerline{\includegraphics[scale=0.22]{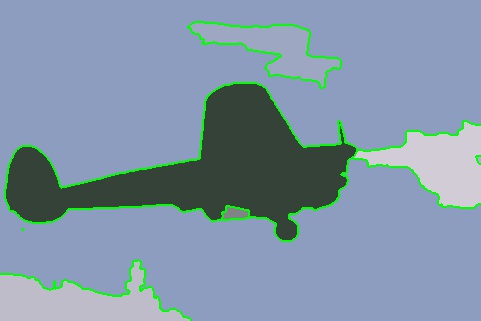}}
\end{minipage}
\begin{minipage}{0.46\linewidth}
\centerline{\includegraphics[scale=0.22]{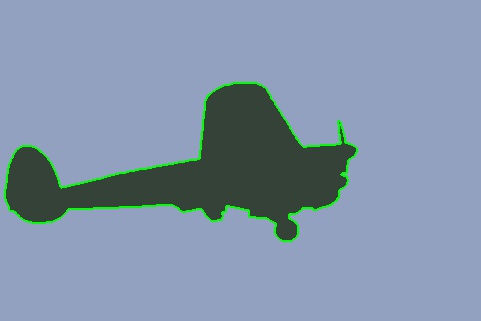}}
\end{minipage}
\\
\begin{minipage}{0.46\linewidth}
\centerline{(c)}
\end{minipage}
\begin{minipage}{0.46\linewidth}
\centerline{(d)}
\end{minipage}
  \caption{(a) Input image; (b) Result of the proposed network trained by using only intra-consistency $\calL_{local}$ within each superpixel; (c) Result of the proposed network; (d) Result of the proposed framework including a graph-based post-processing method.}
\label{fig:teaser}
\end{center} \end{figure}

To overcome these limitations, in this paper, we propose a novel framework that can extract more discriminative and consistent features by utilizing CNNs, image reconstruction, and superpixel segmentation, as illustrated in~\fref{fig:overview}. Unlike previous works, we introduce a feature embedding module (FEM) to replace typical residual blocks in CNNs. The FEM employs a channel attention mechanism and a fused activation function. Furthermore, we propose to explicitly aggregate local and global context information. Lastly, we propose a novel loss function that utilizes both superpixel segmentation and image reconstruction. The loss for image reconstruction is computed using both structural similarity (SSIM) and pixel-level similarity. The loss using superpixels considers both the internal consistency within each superpixel and the inter-similarity/dissimilarity between neighboring superpixels (see~\fref{fig:teaser}). We evaluate the proposed framework on two public benchmark datasets (BSDS dataset~\citep{martin2001database} and PASCAL VOC 2012 dataset~\citep{Everingham2010}) for unsupervised image segmentation. The experimental results demonstrate that our proposed method outperforms the previous state-of-the-art method~\citep{dic2020}.

The contributions of this paper are as follows:
\begin{itemize}
  \item We introduce a pixel-wise clustering network that utilizes a channel-wise attention mechanism and aggregates both local and global contextual information to achieve high accuracy. 
  \item To train the network with only a single image and without requiring any annotations, we propose a novel loss function. This function penalizes clustering pixels with similar features in neighboring superpixels into different clusters. It also penalizes classifying pixels within each superpixel into different clusters. It also uses both multi-scale structural similarity (MS-SSIM) and L2 losses. 
  \item We propose a method for using the statistics of both deep and shallow features to measure the similarity of features between neighboring superpixels. 
  \item We demonstrate the extended value of the proposed method by applying it to unsupervised semantic segmentation. 
\end{itemize}

\section{RELATED WORKS}

\subsection{Unsupervised Image Segmentation}

As various computer vision tasks have achieved high accuracy with CNN-based methods, unsupervised image segmentation methods have also employed CNNs. One of the earliest works that utilized CNNs for unsupervised image segmentation was proposed by \cite{xia2017w}. They introduced the W-Net, which consists of an encoder and a decoder. The network is trained by minimizing image reconstruction loss and normalized cut loss. They also employed post-processing techniques such as conditional random field (CRF) smoothing \citep{krahenbuhl2011efficient} and hierarchical merging \citep{arbelaez2010contour} to improve the segmentation results. Another early work based on CNNs was proposed independently by \cite{Kanezaki2018}. While \cite{xia2017w} utilized an image reconstruction loss, \cite{Kanezaki2018} trained the network using separately extracted superpixels. Both methods designed their networks so that each channel of the output corresponds to the probability of belonging to each cluster. Therefore, pixel-level cluster labels are obtained by finding the maximum value along the channels.

To overcome the limitations associated with losses dependent on superpixel segmentations, \cite{kim2020unsupervised} and \cite{kim2020} independently devised techniques that eliminate the need for superpixels. Instead of relying on a superpixel-based loss, \cite{kim2020unsupervised} employed a spatial continuity loss function that encourages neighboring pixels to be grouped together, while \cite{kim2020} utilized a Mumford-Shah functional-based loss function to train CNNs. \cite{Lin2020} and \cite{dic2020} improved upon the previous works that relied on superpixel segmentations. \cite{Lin2020} proposed a framework that employs both an autoencoder similar to \citep{xia2017w} and superpixel segmentation like \citep{Kanezaki2018}. \cite{dic2020} proposed a trainable clustering module that iteratively updates cluster associations and cluster centers.

In this paper, we propose a CNN-based clustering framework that utilizes both image reconstruction and superpixel segmentation to compute a loss. To extract more discriminative and consistent representations for segmentation than previous works, we replace the typical residual block with the proposed feature embedding module (FEM) and fuse local and global information using explicit multi-scaling. The FEM employs a channel-attention mechanism and a fused activation function. Unlike previous works \citep{xia2017w, Lin2020}, we compute the image reconstruction loss using both patch-wise structural differences and pixel-level differences. For superpixel segmentation, unlike previous methods \citep{Kanezaki2018, dic2020}, we compute a loss based on both inter-similarity/dissimilarity between neighboring superpixels and intra-identity within each superpixel. Compared to the method by \cite{Lin2020}, we propose additional statistics to measure inter-similarity/dissimilarity. Experimental results demonstrate that the proposed method achieves about 10\% relatively higher accuracy than the previous state-of-the-art method.

\subsection{Unsupervised Semantic Segmentation}

Unsupervised image segmentation and unsupervised semantic segmentation serve different purposes and have different requirements. While unsupervised image segmentation clusters pixels within an image into separate regions based on instances, objects, and components, unsupervised semantic segmentation classifies pixels based on semantic classes. Accordingly, the latter usually requires a set of training images (patches) to handle semantic classes while the former can be trained using only a single image.

Despite their differences, we briefly summarize previous works on unsupervised semantic segmentation since both tasks aim to learn how to cluster pixels without ground-truth annotations. \cite{IIC2019} introduced a technique that extracts common representations from the same objects while discarding instance-specific features by employing random transformations and spatial proximity. \cite{Cho2021} proposed a method that uses both geometric and photometric transformations to generate multiple augmented versions of original images. \cite{Van2021} proposed a method that uses object mask proposals and a contrastive loss function. The method first generates object masks and then uses them to train feature embeddings. Most recently, \cite{hamilton2022unsupervised} introduced STEGO, a method that uses a pre-trained and frozen backbone to extract features, and then distills them into discrete semantic labels using contrastive learning.

Since the two tasks share some commonalities, certain approaches can be applied to both. Therefore, following the previous literature by \cite{kim2020unsupervised}, we compare our proposed method to IIC~\citep{IIC2019}. Additionally, we present an extension of the proposed method for unsupervised semantic segmentation by fusing it with STEGO~\citep{hamilton2022unsupervised}.

\begin{figure*}[!t] \begin{center}
\begin{minipage}{0.98\linewidth}
\centerline{\includegraphics[scale=0.71]{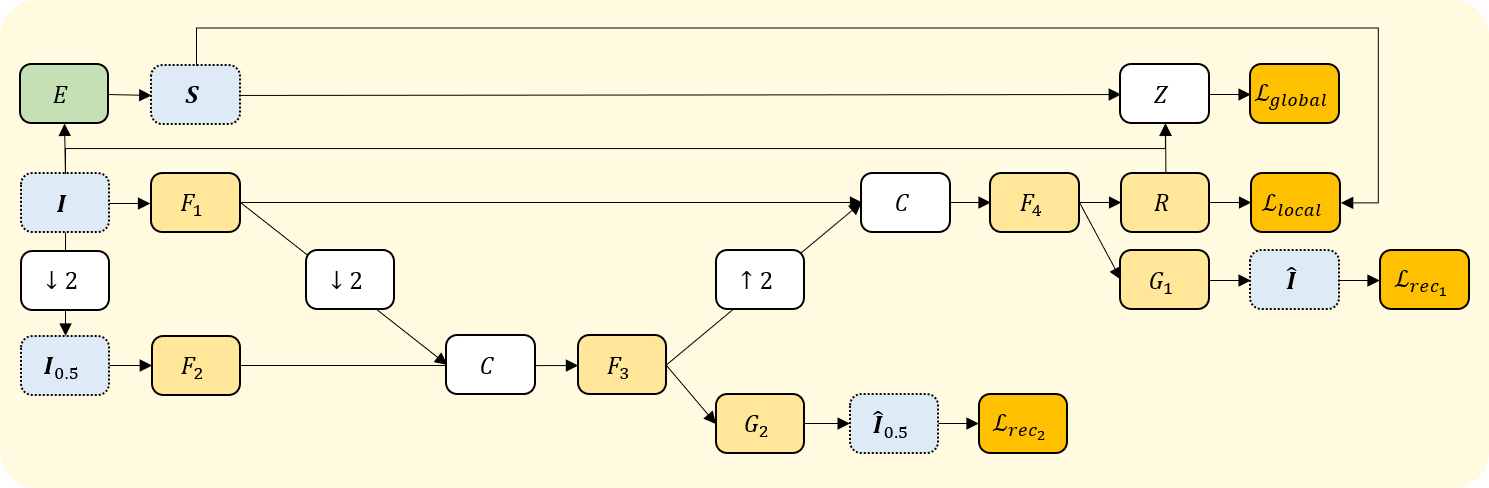}}
\end{minipage}
   \caption{The proposed framework during training. $\mI$, $\mI_{0.5}$, and $\mS$ represent an image, its downsampled image, and extracted superpixels, respectively. $E$ and ($F_1$, $F_2$, $F_3$, $F_4$) denote superpixel segmentation and four feature embedding modules, respectively. ($G_1$, $G_2$) and ($\hat{\mI}$, $\hat{\mI}_{0.5}$) represent two image reconstruction modules and reconstructed images, respectively. $Z$, $R$, $C$, $\downarrow2$, and $\uparrow2$ represent a feature statistics computing module, a cluster prediction module, concatenation, downsampling by a factor of 2, and upsampling by a factor of 2, respectively. $\calL_{local}$ and $\calL_{global}$ denote loss terms within a superpixel and between neighboring superpixels, respectively. $\calL_{rec}$ denotes an image reconstruction loss.}
\label{fig:framework}
\end{center}\end{figure*}

\section{PROPOSED METHOD}
\label{sec:method}

\subsection{Pixel-level Clustering Network}
\label{sec:fem}
The proposed framework is designed to achieve accurate unsupervised image segmentation by utilizing feature statistics, fusing local and global context features, and employing attention mechanisms, image reconstruction, explicit multi-scaling, and superpixel segmentation. An overview of the proposed framework is shown in~\fref{fig:framework}. 

To robustly determine the merging or separating of clusters, feature statistics are computed for each superpixel using extracted features from CNNs. These statistics are then utilized to compare neighboring superpixels. The feature statistics computing module, superpixel segmentation algorithm, and extracted superpixels are denoted as $Z$, $E$, and $\mS$, respectively, in~\fref{fig:framework}. Further details of the feature statistics computing module are described in~\sref{sec:loss}. 

The fusion of local and global context features is essential to achieve accurate segmentation by considering both adjacent regions and global circumstances. To accomplish this, we combine feature maps extracted at input resolution with those extracted at half of the input resolution. 
We perform feature extraction using four feature embedding modules ($F_1$, $F_2$, $F_3$, $F_4$ in~\fref{fig:framework}). Given an input image $\mI$, $F_1$ extracts feature maps at input resolution that correspond to relatively local features. 
We use bicubic interpolation-based downsampling instead of a strided convolution to explicitly downsample $\mI$ by 2 and extract feature maps from the downsampled image $\mI_{0.5}$. The extracted maps contain relatively global features. Then, we downsample the output of $F_1$ by 2 using max-pooling and concatenate ($C$) it with the output of $F_2$. The outputs of $F_1$ and $F_2$ complement each other as the former provides more local information, and the latter contains more global representation. $F_3$ takes the concatenated maps and extracts representations that correspond to global context information. We concatenate the output of $F_1$ and the upsampled output of $F_3$ and process them using $F_4$.

The feature embedding module (FEM) combines the attention mechanism with the structure of a residual unit to achieve accurate segmentation. The attention mechanism enables the neurons to focus on significant features and suppress irrelevant representations, while the residual structure ensures stable training. Together with the explicit fusion of local and global features, these contribute to learning and extracting more robust and meaningful features for unsupervised image segmentation.

Superpixel segmentation is used to compute a loss for training the network. First, an image $\mI$ is passed through a superpixel segmentation algorithm $E$ to obtain a set of superpixels $\mS$. Assuming that the superpixel segmentation is accurate, a loss is then computed to ensure that the pixels within a superpixel are clustered together. As superpixels are not directly used to segment an image, minor inaccuracy in superpixel segmentation is acceptable for computing a loss to train the network. As mentioned earlier, the extracted superpixels are also used to compute feature statistics to compare neighboring superpixels. 

The image reconstruction modules ($G_1$, $G_2$) guide the clustering network to encode sufficient information for robust clustering. They force the network to consider the overall content of the image at intermediate stages, rather than making clustering predictions at early layers. Each module consists of one $1 \times 1$ convolution layer, which reduces the number of channels to that of color channels. We empirically demonstrate that the image reconstruction modules improve accuracy.

The followings describe more details about the training of the clustering framework. Firstly, given an image $\mI$, the superpixel segmentation algorithm $E$ is applied to obtain the superpixels $\mS$. Since the superpixel segmentation is not dependent on the parameters in CNNs, the superpixel extraction is performed only once per image. While any superpixel segmentation algorithm can be used, we employ Multiscale Combinatorial Grouping (MCG)~\citep{arbelaez2014multiscale}.

The clustering network is also given the image $\mI$. Firstly, the image is downsampled by 2 using bicubic interpolation. Then, $F_1$ processes the original image, and $F_2$ processes the downsampled image. The output of $F_1$ is downsampled by 2 using max-pooling and concatenated with the output of $F_2$. The resulting concatenated feature maps are then processed by $F_3$ and upsampled by a factor of 2 using transposed convolution. Finally, the output of $F_1$ and that of $F_3$ are concatenated and processed by $F_4$. The outputs of $F_1$, $F_2$, $F_3$, and $F_4$ have 64, 64, 128, and 128  channels, respectively. 

\begin{figure*}[!t] \begin{center}
\begin{minipage}{0.98\linewidth}
\centerline{\includegraphics[scale=0.7]{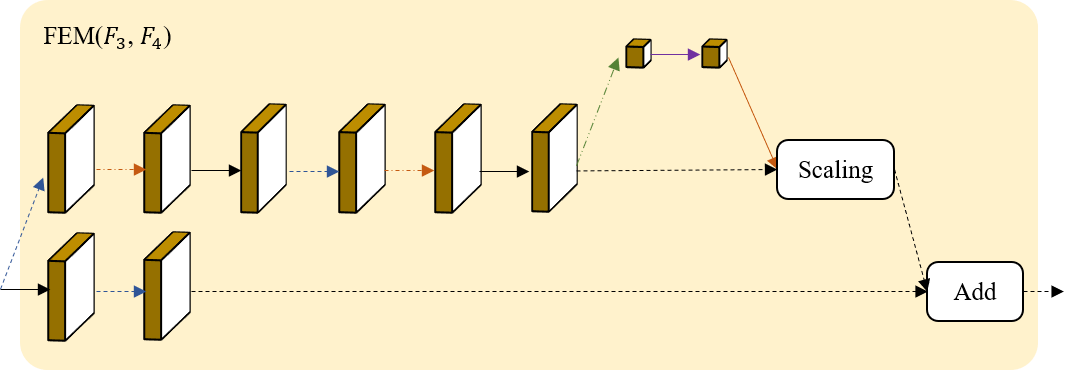}}
\end{minipage}
   \caption{Architecture of feature embedding module (FEM) for $F_3$ and $F_4$. Each line and each 3D volume correspond to an operation and a feature map(vector), respectively. Black solid and dashed lines represent a convolution and an information flow, respectively. Blue dashed and orange dot-dash lines denote batch normalization and ReLU+$tanh$ activation function, respectively. Green dot-dot-dash, purple solid, and orange solid lines represent global average pooling, a 1D convolution layer, and a sigmoid function, respectively.}
\label{fig:fem}
\end{center}\end{figure*}

\begin{figure}[!t] \begin{center}
\begin{minipage}{0.98\linewidth}
\centerline{\includegraphics[scale=0.56]{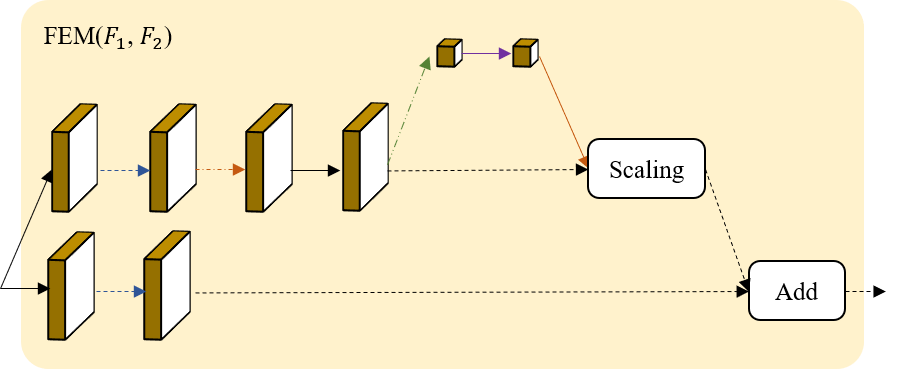}}
\end{minipage}
   \caption{Architecture of feature embedding module (FEM) for $F_1$ and $F_2$. Notations are the same as~\fref{fig:fem}.}
\label{fig:fem2}
\end{center}\end{figure}

The FEM has a structure similar to that of a residual block~\citep{he2016identity, he2016resnet}. In~\fref{fig:fem}, the bottom connection corresponds to a shortcut connection, consisting of a $3 \times 3$ convolution layer and batch normalization~\citep{ioffe2015batch}. The middle connection is a typical stacked convolution module that contains two stacked blocks, each with batch normalization, an activation function, and a $3 \times 3$ convolution layer. We use a weighted summation of ReLU and $tanh$ activation functions~\citep{li2020combine} for the activation function. After the two stacked blocks, we apply an attention mechanism similar to the Efficient Channel Attention (ECA) block~\citep{ecanet2020}. The attention mechanism scales the output of the stacked block by a predicted significance for each channel, which is predicted by the layers presented at the top in~\fref{fig:fem}. These layers consist of global average pooling, a 1D convolution layer, and a sigmoid activation function. Finally, the output of the FEM is obtained by adding the output of the shortcut connection and the significance-scaled output of the stacked convolution module. Figures~\ref{fig:fem} and~\ref{fig:fem2} show the structures of ($F_3$, $F_4$) and ($F_1$, $F_2$), respectively. The main differences are the absence of batch normalization and an activation function at the front of the middle connection in $F_1$ and $F_2$. Since $F_1$ and $F_2$ process images instead of feature maps, these modules do not apply them at the front.

The output of $F_4$ is used by $R$ to predict the probability of each pixel belonging to each cluster. To achieve this, a $1 \times 1$ convolution layer and batch normalization are employed. The cluster label for each pixel is then determined by selecting the cluster with the highest probability.

\subsection{Loss Function}
\label{sec:loss}
As pixels that are close in distance and have similar features are likely to belong to the same cluster, we use the features extracted using CNNs and the spatial distance to calculate a loss. Additionally, because superpixel segmentation algorithms have been extensively studied~\citep{slic2012, arbelaez2014multiscale}, we use one of them as a guide to compute a loss. 

The proposed loss function consists of three terms. The first term aims to ensure that pixels within each superpixel belong to the same cluster. The second term encourages neighboring superpixels to belong to the same cluster if their corresponding features are similar. The third term encodes image information in the clustering network.

To train the clustering network, we first extract superpixels from the input image $\mI$. Specifically, we use Multiscale Combinatorial Grouping (MCG)~\citep{arbelaez2014multiscale} to extract $K$ superpixels $\{S_k\}_{k=1}^K$. Then, we use the extracted superpixels to compute the loss. At each iteration, the proposed model clusters the pixels in $\mI$ into multiple segments (clusters). Assuming that superpixel segmentation provides reliable results, the pixels within each superpixel should belong to the same cluster. Therefore, we find the most frequent cluster label for each superpixel and consider it as the pseudo ground-truth. We then compute the pixel-wise cross-entropy loss by comparing the output of the proposed model with the pseudo ground-truth.

In more detail, given the result of superpixel segmentation and that of the proposed model, for each superpixel $S_k$, the most frequent cluster $c^m_k$ is found as follows:
\begin{equation}
	c^m_k = c^j \text{ where } j=\argmax{i}{|c^i_k|}
\label{eq:cmk}
\end{equation}
where $|c^i_k|$ denotes the number of pixels that belong to the cluster $c^i$ and are in the superpixel $S_k$. By utilizing the superpixel-wise most frequent cluster label $c^m_k$ found by~\eref{eq:cmk}, we can construct $\mC^m_{n,q}$ that contains the pseudo ground-truth for each pixel. Then, the cross-entropy loss is computed as follows:
\begin{equation}
	\calL_{local}=-\sum_{n=1}^{N} \sum_{q=1}^{Q} \mathbbm{1}(\mC^m_{n,q},q) \ln{\mP_{n,q}}
\label{eq:loss_local}
\end{equation}
where $Q$ denotes the number of channels of output, which corresponds to the maximum number of clusters. $\mP_{n,q}$ represents the output of the model at $(n,q)$. $\mathbbm{1}(\mC^m_{n,q},q)$ denotes an indicator function that returns one if $\mC^m_{n,q}$ and $q$ are the same and zero otherwise. $N$ and $n$ represent the total number of pixels in $\mI$ and the index for each pixel, respectively. 

As $\calL_{local}$ considers pixels in each superpixel separately, we also compute another loss term $\calL_{global}$ to ensure that pixels are clustered together if they belong to neighboring superpixels and have similar features. The loss is computed by utilizing the feature statistics computing module. 

The extracted superpixels are first used to construct a graph in which each superpixel corresponds to a node. The features ($\vv_k$, $\ve_k$) of each node are then calculated using the output $\mP$ of the clustering network and the input image $\mI$, respectively. Specifically, the deep feature $\vv_k \in \mathbb{R}^{Q}$ is computed by summing the mean and standard deviation of the features $\mP$ of the pixels that belong to the corresponding superpixel. The mean and standard deviation of the features are calculated to consider their distribution.
\begin{equation}
\begin{split}
	\vec{\mu}_k &= \frac{1}{|S_k|} \sum_{i \in S_k} \mP_i, \\
	\vec{\sigma}_k &= \sqrt{\frac{1}{|S_k|} \sum_{i \in S_k}(\mP_i - \vec{\mu}_k)^2}, \\
	\vv_{k} &= \vec{\mu}_k + \vec{\sigma}_k
\end{split}
\end{equation}
where $|S_k|$ denotes the number of pixels in the superpixel $S_k$. The shallow feature $\ve_k \in \mathbb{R}^{3}$ is computed using the same method as $\vv_k$, with $\mP$ in the above equation is replaced by $\mI$. 

Given the nodes corresponding to superpixels, the loss is computed by taking into account the connectivity between nodes and the similarity between their features. For connectivity, nodes are connected by an edge if the corresponding superpixels are neighboring. Specifically, if any two different superpixels share a common border, the corresponding nodes are connected by an edge. For similarity, an affinity matrix $\mA \in \mathbb{R}^{K \times K}$ is computed as follows:
\begin{equation}
	\mA_{i,j}=\left\{
	\begin{array}{l l}
	\text{exp} \Big(-\frac{\norm{\vv_i-\vv_j}^2_2}{\alpha_1}-\frac{\norm{\ve_i-\ve_j}^2_2}{\alpha_2}\Big), \; \text{ if } \mB_{i,j}=1 \\
	0, \; \text{ otherwise } \\ 
	\end{array} \right.
\label{eq:affinity}	
\end{equation}
where $\mB_{i,j}$ is one if $S_i$ and $S_j$ are different and neighboring superpixels, and is zero otherwise. $\alpha_1$ and $\alpha_2$ are hyperparameters. 

Then, $\calL_{global}$ is computed as follows:
\begin{equation}
	\calL_{global}=\frac{1}{|\mA|} tr(\mH^\text{T} \mA (\textbf{1}-\mH)) 
\label{eq:loss_global}
\end{equation}
where $|\mA|$ denotes the summation of all elements in $\mA$ computed by~\eref{eq:affinity}. $tr(\cdot)$ represents a trace operation. $\mH \in \mathbb{R}^{K \times Q}$ contains the probability of the superpixel at each row belonging to the cluster at each column. Given $\mP$ from the proposed model, a softmax function is firstly applied for normalization. Then, for each superpixel, the outputs of the softmax function are averaged to obtain each row of $\mH$.

Image reconstruction loss $\calL_{rec}$ consists of two MS-SSIM+$\ell_2$ losses introduced by~\cite{Zhao2017} as follows: 
\begin{equation}
	\calL_{rec} = \calL_{ms\text{-}ssim+\ell_2}(\mI, \hat{\mI}) + \calL_{ms\text{-}ssim+\ell_2}(\mI_{0.5}, \hat{\mI}_{0.5}) 
\label{eq:loss_rec}
\end{equation}
where $\hat{\mI}$ and $\hat{\mI}_{0.5}$ denote the reconstructed images at the original resolution and the half resolution, respectively. The two terms on the right side are denoted by $\calL_{rec_1}$ and $\calL_{rec_2}$, respectively, in~\fref{fig:framework}. $\calL_{ms\text{-}ssim+\ell_2}(\cdot, \cdot)$ is described in~\eref{eq:msssim_l2}.

MS-SSIM+$\ell_2$ loss is a weighted summation of a MS-SSIM loss and a L2 loss. $\calL_{ms\text{-}ssim+\ell_2}(\mI, \hat{\mI})$ is computed as follows:
\begin{equation}
\begin{split}
	& \calL_{ms\text{-}ssim+\ell_2}(\mI, \hat{\mI})\\ 
	& =\eta \calL_{ms\text{-}ssim}(\mI, \hat{\mI}) + (1-\eta) G_{\sigma^M_G}* \calL_{\ell_2}(\mI, \hat{\mI}) 
\end{split}
\label{eq:msssim_l2}
\end{equation}
where $\eta$ is a weighting coefficient to balance between the MS-SSIM loss and the L2 loss. $G_{\sigma^M_G}$ represents Gaussian filters with the standard deviations of $\sigma^M_G$ for varying scales $M$. $\sigma^{1}_G$, $\sigma^{2}_G$, $\sigma^{3}_G$, $\sigma^{4}_G$, and $\sigma^{5}_G$ are $0.5$, $1$, $2$, $4$, and $8$, respectively. $*$ denotes a convolution operation. $\calL_{ms\text{-}ssim}(\mI, \hat{\mI})$ is computed as follows: 
\begin{equation}
	\calL_{ms\text{-}ssim}(\mI, \hat{\mI})=1-\text{MS-SSIM}(\mI, \hat{\mI}) 
\end{equation}
where MS-SSIM is computed by the method in~\citep{msssim2003, msssim2004}. We refer readers to~\citep{Zhao2017} for more details. 

Finally, the total loss $\calL$ is computed by a weighted sum of $\calL_{local}$, $\calL_{global}$, and $\calL_{rec}$ as follows:
\begin{equation}
	\calL = \calL_{local} + \gamma_1 \calL_{global} + \gamma_2 \calL_{rec} 
\label{eq:total_loss}
\end{equation}
where $\gamma_1$ and $\gamma_2$ are weighting coefficients to balance three loss terms. 
$\calL_{local}$, $\calL_{global}$, and $\calL_{rec}$ are computed by Eqs.~(\ref{eq:loss_local}),~(\ref{eq:loss_global}), and~(\ref{eq:loss_rec}), respectively.

\begin{algorithm}[!t]
\caption{Algorithm for training process}

\begin{algorithmic}[1]
\State \textbf{Input:} Input image $\mI \in \mathbb{R}^{H \times W \times 3 }$ 
\State \textbf{Output:} Cluster label map $\mI^* \in \mathbb{Z}^{H \times W }$  
\State Initialize the hyperparameters $\alpha_1, \alpha_2, \gamma_1, \gamma_2, K$
\State Initialize the network parameters $W$
\State Extract superpixels $\{S_k\}^K_{k=1}$

\For{$t=1$ to $T$}
 \State {Extract $\mP$ from $\mI$ using current $W$ in the network.}
 \State {Obtain $\mI^*$ by applying $\text{argmax}$ to $\mP$.}
 \State {Compute loss $\calL$ using $\mP$, $\mI^*$, $\mI$, $\hat{\mI}$, and $S_k$ by~\eref{eq:total_loss}}.
 \State {Update $W$ by minimizing $L$.}
\EndFor

\end{algorithmic}
\end{algorithm}

\subsection{Training Process}
\label{sec:training}
The training process is outlined in Algorithm 1. Given an input image $\mI$, we begin by extracting superpixels $S_k$. At each iteration $t$, we forward-propagate $\mI$ through the clustering network with the current parameters $\mW$. The pixel-wise clustering result $\mI^*$ is then obtained by finding the arguments of the maxima. The loss $\calL$ in~\eref{eq:total_loss} is computed by utilizing the clustering result $\mI^*$, the network's output $\mP$, reconstructed images ($\hat{\mI}$, $\hat{\mI}_{0.5}$), and superpixels $S_k$ to update the parameters in the network through back-propagation. The optimization method used is stochastic gradient descent with momentum. The iteration is repeated for a predetermined number of iterations.


\begin{table*}[!t]
\small
\begin{center}
\begin{minipage}{0.9\linewidth}
\caption{Quantitative results on the BSDS300 dataset~\citep{martin2001database}.}
\label{tab:result_300}
\begin{tabular}{>{\centering}m{0.4\textwidth}|>{\centering}m{0.105\textwidth}|>{\centering}m{0.105\textwidth}|>{\centering}m{0.105\textwidth}|>{\centering\arraybackslash}m{0.105\textwidth}} 
\hline
Methods & PRI & VoI & GCE & BDE \\ 
\hline\hline
FH~\citep{graph2004}& 0.714 & 3.395 & 0.175 & 16.67  \\ 
NCuts~\citep{868688} & 0.724 & 2.906 & 0.223 & 17.15  \\ 
NTP~\citep{wang2008normalized}& 0.752 & 2.495 & 0.237 & 16.30  \\ 
MNCut~\citep{cour2005spectral}& 0.756 & 2.44 & 0.193 & 15.10  \\ 
KM~\citep{salah2010multiregion}& 0.765 & 2.41 & - & -  \\ 
JSEG~\citep{deng2001unsupervised}& 0.776 & 1.822 & 0.199 & 14.40  \\ 
SDTV~\citep{donoser2009saliency}& 0.776 & 1.817 & 0.177 & 16.24  \\ 
Mean-Shift~\citep{1000236} & 0.796 & 1.973 & 0.189 & 14.41  \\ 
TBES~\citep{mobahi2011segmentation}& 0.80 & 1.76 & - & -  \\ 
CCP~\citep{fu2015robust}& 0.801 & 2.472 & \textbf{0.127} & 11.29  \\ 
MLSS~\citep{kim2012learning}& 0.815 & 1.855 & 0.181 & 12.21  \\  
gPb-owt-ucm \citep{arbelaez2009contours} & 0.81 & \underline{1.68} & - & -  \\ 
W-Net~\citep{xia2017w}& 0.81 & 1.71 & - & -  \\ 
SAS~\citep{li2012segmentation}& 0.832 & 1.685 & 0.178 & 11.29  \\ 
DIC~\citep{dic2020} & \underline{0.841} & 1.749 & \underline{0.139} & \underline{10.18}  \\ 
\hline
Proposed & \textbf{0.867} & \textbf{1.399} & 0.163 & \textbf{8.832}  \\ 
 \hline
\end{tabular}
\end{minipage}
\end{center}
\end{table*}

\subsection{Post-processing}
\label{sec:postprocessing}
As the proposed clustering network is trained using superpixels along with others, the network performs well in segmenting relatively small regions. However, the network may struggle with accurately segmenting large regions, such as the background. To address this issue, a post-processing method is proposed, which involves constructing an undirected graph and using graph cuts to obtain the final segmentation result. Each cluster is represented as a node in the undirected graph, and edge weights are computed using image gradients. Finally, the edges are cut based on their weights to obtain the final segmentation.

Given the pixel-wise clustering result $\mI^*$, an undirected graph is constructed by considering each segment $\hat{S}$ in $\mI^*$ as a vertex. To compute edge weights, the input image $\mI$ is first converted to the image $\tilde{\mI}$ in the CIELAB color space. Gradients ($\mI^x$, $\mI^y$) are then computed along the x- and y-axes using $\tilde{\mI}$.
\begin{equation}
\begin{split}
	\mI^x = | \nabla_x \tilde{\mI} |, \quad
	\mI^y = | \nabla_y \tilde{\mI} |
\end{split}
\label{eq:Ix_Iy}
\end{equation}
where $\nabla_x$ and $\nabla_y$ denote a gradient operation along x- and y-axes, respectively. 
Then, edge weights are computed as follows: 
\begin{equation}
\begin{split}
	\Delta_{i,j} = \left\{
	\begin{array}{l l}
	\|g^x_{i,j} + g^y_{i,j} \|_1, \; \text{ if } \hat{\mB}_{i,j}=1 \\
	\infty, \; \text{ otherwise } \\ 
	\end{array} \right.
\end{split}
\end{equation}
where $g^x_{i,j}$ and $g^y_{i,j}$ are computed by~\eref{eq:gx_gy}. $\hat{\mB}_{i,j}$ is one if $\hat{S}_i$ and $\hat{S}_j$ are different and neighboring segments, and is zero otherwise. $g^x_{i,j}$ and $g^y_{i,j}$ are the average of the absolute difference in the CIELAB color space between $\hat{S}_i$ and $\hat{S}_j$ along x- and y-axes, respectively. They are computed as follows: 
\begin{equation}
\begin{split}
	g^x_{i,j} = \frac{1}{|\tilde{S}^x_{i,j}|} \sum_{(m,n) \in \tilde{S}^x_{i,j}}  \mI^x_{m,n}	\\
	g^y_{i,j} = \frac{1}{|\tilde{S}^y_{i,j}|} \sum_{(m,n) \in \tilde{S}^y_{i,j}}  \mI^y_{m,n}	\\
\end{split}
\label{eq:gx_gy}
\end{equation}
where $\tilde{S}^x_{i,j}$ consists of the pixels at the boundaries between $\hat{S}_i$ and $\hat{S}_j$ along x-axis. Hence, $(m,n) \in \tilde{S}^x_{i,j}$ belongs to $\hat{S}_i$ and its neighboring pixel $(m,n+1)$ belong to $\hat{S}_j$. Similarly, $\tilde{S}^y_{i,j}$ consists of the pixels at the boundaries between $\hat{S}_i$ and $\hat{S}_j$ along y-axis. Accordingly, $(m,n) \in \tilde{S}^y_{i,j}$ belongs to $\hat{S}_i$ and its neighboring pixel $(m+1,n)$ belong to $\hat{S}_j$. $\mI^x_{m,n}$ and $\mI^y_{m,n}$ are calculated by~\eref{eq:Ix_Iy}.

The final segmentation result is obtained by using the graph to cut the edges with high weights. This graph-cut process involves comparing the edge weights to a predetermined threshold, and each connected component in the resulting graph forms a segment.

\section{EXPERIMENTS AND RESULTS}
\label{sec:result}
\subsection{Experimental Setting}
We demonstrate the effectiveness of the proposed framework using the Berkeley Segmentation Data Set (BSDS300 and BSDS500)~\citep{martin2001database, arbelaez2010contour} and the PASCAL VOC 2012 dataset~\citep{Everingham2010}. 
The BSDS500 dataset contains 500 images, which are divided into 200 images for training, 100 for validation, and 200 for testing. The BSDS300 dataset includes only the training and validation splits of the BSDS500 dataset. For evaluation, multiple human annotators label each image in the BSDS dataset. As unsupervised image segmentation methods predict segmentation results using only a single image, they typically do not consider training/validation/test splits separately. Following previous works~\citep{Kanezaki2018, dic2020, kim2020unsupervised}, we train and evaluate the proposed framework using each image in the datasets. For the PASCAL VOC dataset, object category labels are ignored following~\citep{kim2020unsupervised}. Then, the mean Intersection over Union (mIoU) is computed by comparing the segments of the ground truth and those of the predicted results.

Considering hyperparameters, the loss terms were weighted using $\gamma_1=10^{-5}$ and $\gamma_2=0.1$. The affinity matrix was computed using $\alpha_1=200$ and $\alpha_2=400$. The output of the ReLU+$tanh$ activation function was obtained by the weighted summation of the ReLU function and the $tanh$ function, where the weights were 1 and 0.4, respectively. For the stochastic gradient descent optimization, the maximum iteration $T$, learning rate, and momentum were selected as 150, 0.05, and 0.9, respectively.

Following previous works~\citep{dic2020,li2012segmentation}, we utilized optimal image scale (OIS). The proposed framework was applied to each image using six different numbers of superpixels. Among the six results, the best one was used for evaluation. The number $K$ of superpixels were 50, 100, 150, 200, 250, and 300. Please note that superpixels were only employed to train the clustering network.

\begin{table*}[!t]
\small
\begin{center}
\begin{minipage}{0.9\linewidth}
\caption{Quantitative results on the BSDS500 dataset~\citep{martin2001database}.}
\label{tab:result_500}
\begin{tabular}{>{\centering}m{0.4\textwidth}|>{\centering}m{0.15\textwidth}|>{\centering}m{0.15\textwidth}|>{\centering\arraybackslash}m{0.15\textwidth}} 
 \hline
Methods & SC & PRI & VoI  \\ 
\hline\hline
Backprop~\citep{Kanezaki2018} & 0.50 & 0.77 & 2.15 \\ 
NCuts~\citep{868688}& 0.53 & 0.80 & 1.89 \\ 
CAE-TVL~\citep{wang2017unsupervised} & 0.56 & 0.82 & 2.02 \\ 
Mean-Shift~\citep{1000236} & 0.58 & 0.81 & 1.64 \\ 
MLSS~\citep{kim2012learning} & 0.60 & 0.84 & 1.59 \\ 
DSC~\citep{Lin2020}& 0.60 & 0.83 & 1.62 \\ 
W-Net~\citep{xia2017w}& 0.62 & 0.84 & 1.60 \\ 
SF~\citep{dollar2013structured} & 0.65 & 0.851 & 1.43 \\ 
gPb-owt-ucm \citep{arbelaez2009contours}& 0.65 & 0.862 & \underline{1.41} \\ 
DIC~\citep{dic2020} & \underline{0.66} & \underline{0.864} & 1.63 \\ 
\hline
Proposed & \textbf{0.712} & \textbf{0.894} & \textbf{1.305} \\ 
\hline
\end{tabular}
\end{minipage}
\end{center}
\end{table*}

\begin{table*}[!t]
\small
\begin{center}
\begin{minipage}{0.9\linewidth}
\caption{Quantitative results on the PASCAL VOC 2012 dataset~\citep{Everingham2010}.}
\label{tab:result_pascal}
\renewcommand{\arraystretch}{1.1} 
\begin{tabular}{>{\centering}m{0.6\textwidth}|>{\centering\arraybackslash}m{0.3\textwidth}} 
 \hline
Methods & mIoU   \\ 
\hline\hline
$k$-means clustering ($k=2$)  & 0.3166   \\ 
$k$-means clustering ($k=17$) & 0.2383   \\ 
FH~\citep{graph2004} ($\tau=100$) & 0.2682   \\ 
FH~\citep{graph2004} ($\tau=500$) & \underline{0.3647}   \\ 
IIC~\citep{IIC2019} ($k=2$) & 0.2729   \\ 
IIC~\citep{IIC2019} ($k=20$) & 0.2005  \\ 
Backprop~\citep{Kanezaki2018} & 0.3082 \\
DFC~\citep{kim2020unsupervised} ($\mu=5$) & 0.3520 \\ 
\hline
Proposed & \textbf{0.4103}   \\ 
\hline
\end{tabular}
\end{minipage}
\end{center}
\end{table*}

\begin{figure*}[!t] 
\begin{center}
\begin{minipage}{0.02\linewidth}
\centerline{(a)}
\end{minipage}
\begin{minipage}{0.16\linewidth}
\centerline{\includegraphics[scale=0.17]{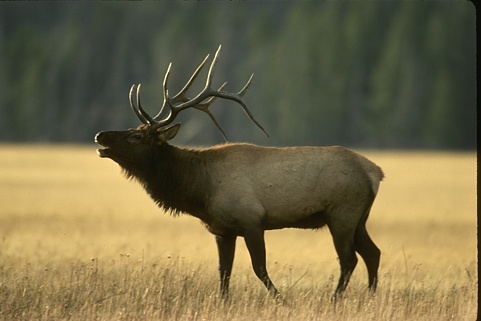}}
\end{minipage}
\begin{minipage}{0.16\linewidth}
\centerline{\includegraphics[scale=0.17]{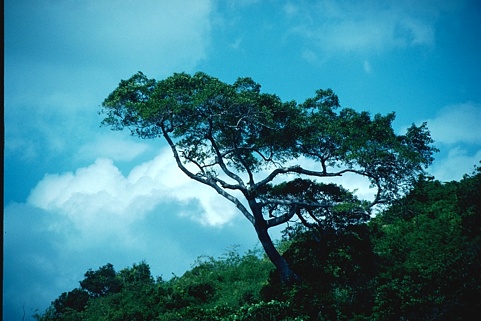}}
\end{minipage}
\begin{minipage}{0.16\linewidth}
\centerline{\includegraphics[scale=0.17]{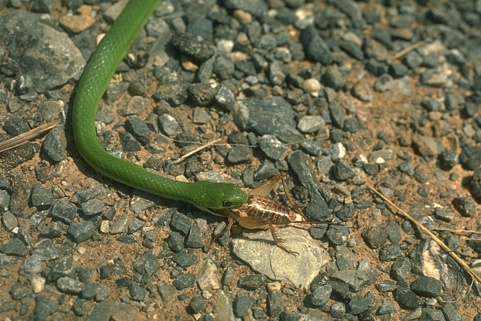}}
\end{minipage}
\begin{minipage}{0.16\linewidth}
\centerline{\includegraphics[scale=0.17]{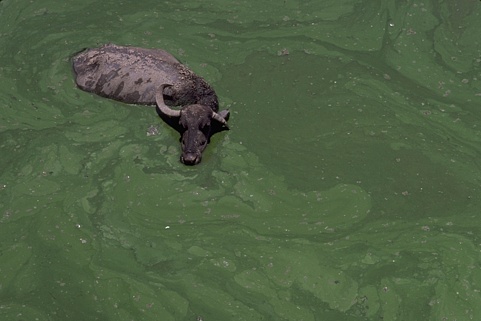}}
\end{minipage}
\begin{minipage}{0.072\linewidth}
\centerline{\includegraphics[scale=0.11322]{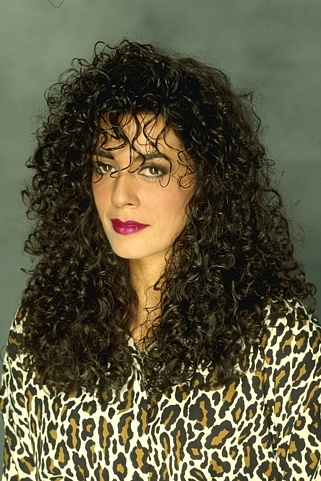}}
\end{minipage}
\begin{minipage}{0.072\linewidth}
\centerline{\includegraphics[scale=0.11322]{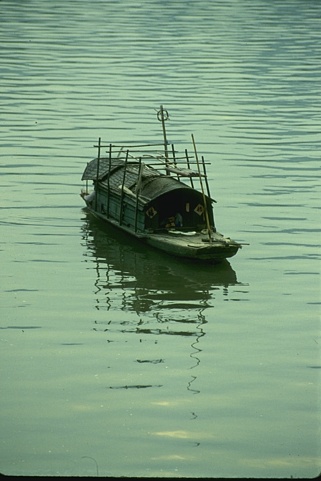}}
\end{minipage}
\begin{minipage}{0.072\linewidth}
\centerline{\includegraphics[scale=0.11322]{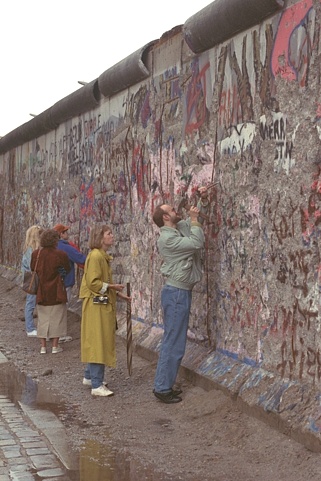}}
\end{minipage}
\begin{minipage}{0.072\linewidth}
\centerline{\includegraphics[scale=0.11322]{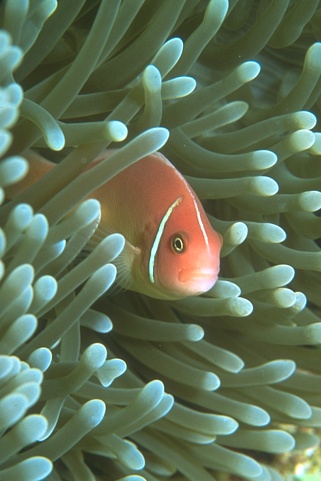}}
\end{minipage}
\\

\vspace{0.1cm}
\begin{minipage}{0.02\linewidth}
\centerline{(b)}
\end{minipage}
\begin{minipage}{0.16\linewidth}
\centerline{\includegraphics[scale=0.17]{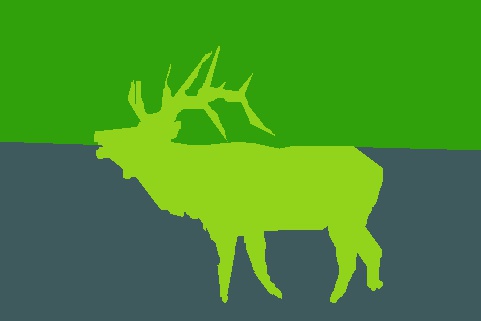}}
\end{minipage}
\begin{minipage}{0.16\linewidth}
\centerline{\includegraphics[scale=0.17]{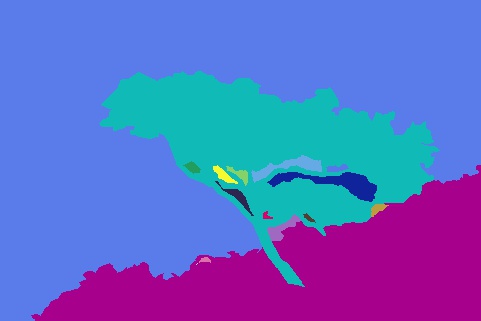}}
\end{minipage}
\begin{minipage}{0.16\linewidth}
\centerline{\includegraphics[scale=0.17]{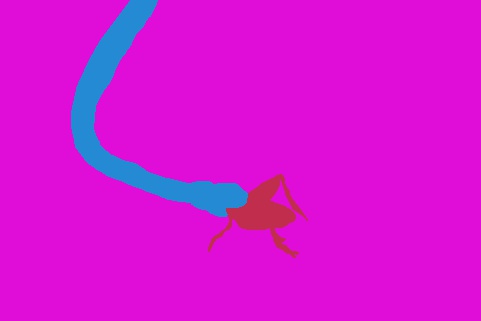}}
\end{minipage}
\begin{minipage}{0.16\linewidth}
\centerline{\includegraphics[scale=0.17]{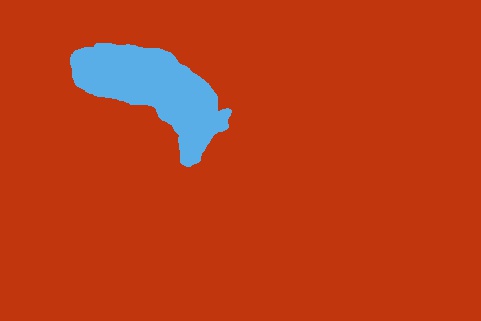}}
\end{minipage}
\begin{minipage}{0.072\linewidth}
\centerline{\includegraphics[scale=0.11322]{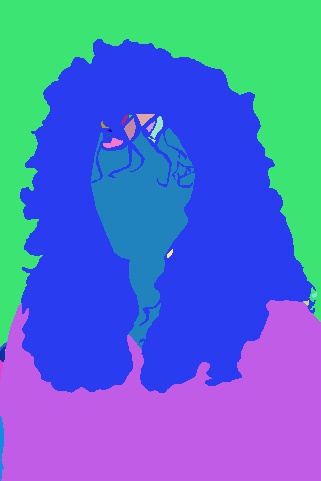}}
\end{minipage}
\begin{minipage}{0.072\linewidth}
\centerline{\includegraphics[scale=0.11322]{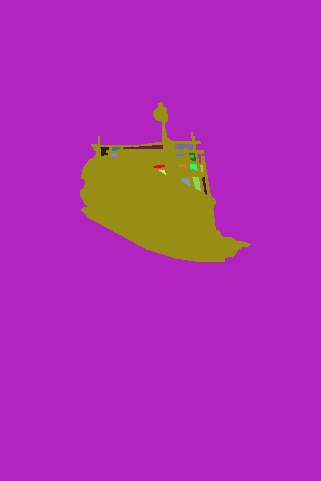}}
\end{minipage}
\begin{minipage}{0.072\linewidth}
\centerline{\includegraphics[scale=0.11322]{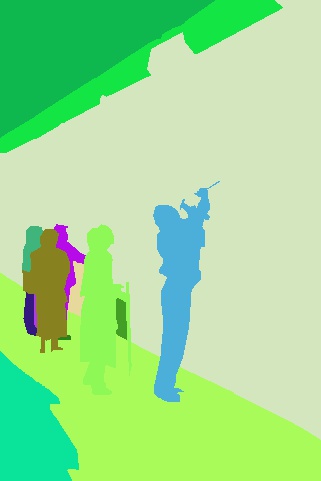}}
\end{minipage}
\begin{minipage}{0.072\linewidth}
\centerline{\includegraphics[scale=0.11322]{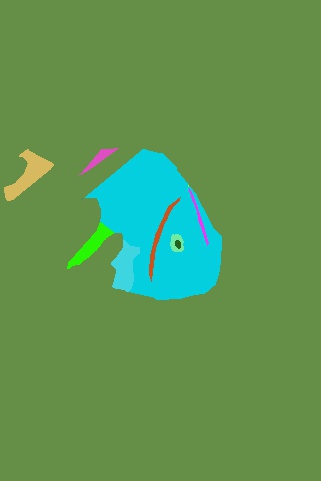}}
\end{minipage}
\\

\vspace{0.1cm}
\begin{minipage}{0.02\linewidth}
\centerline{(c)}
\end{minipage}
\begin{minipage}{0.16\linewidth}
\centerline{\includegraphics[scale=0.17]{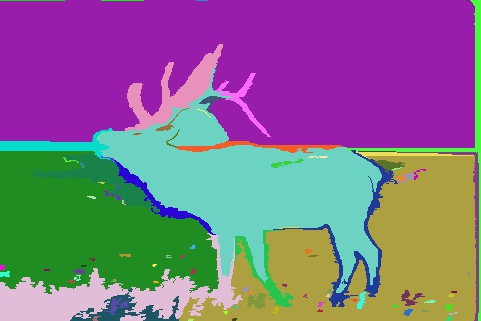}}
\end{minipage}
\begin{minipage}{0.16\linewidth}
\centerline{\includegraphics[scale=0.17]{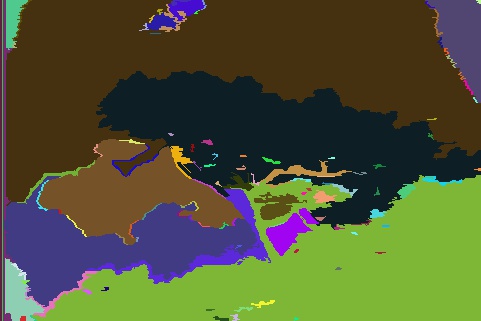}}
\end{minipage}
\begin{minipage}{0.16\linewidth}
\centerline{\includegraphics[scale=0.17]{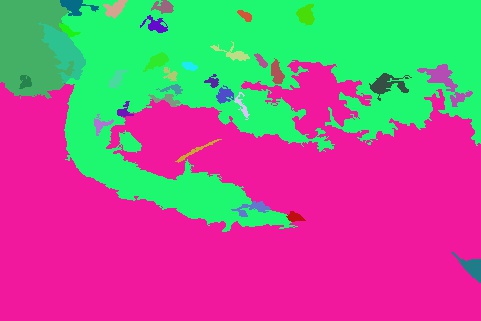}}
\end{minipage}
\begin{minipage}{0.16\linewidth}
\centerline{\includegraphics[scale=0.17]{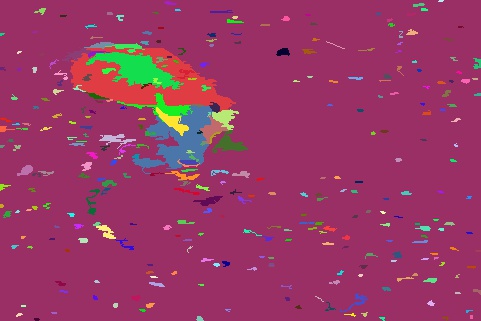}}
\end{minipage}
\begin{minipage}{0.072\linewidth}
\centerline{\includegraphics[scale=0.11322]{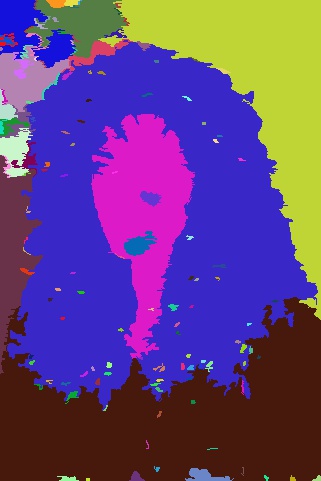}}
\end{minipage}
\begin{minipage}{0.072\linewidth}
\centerline{\includegraphics[scale=0.11322]{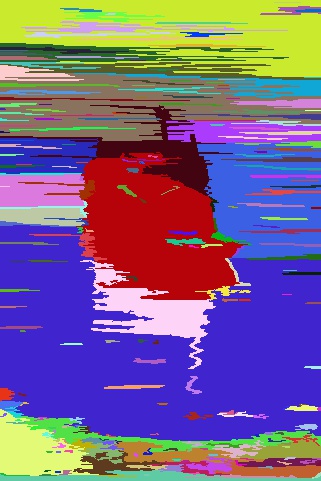}}
\end{minipage}
\begin{minipage}{0.072\linewidth}
\centerline{\includegraphics[scale=0.11322]{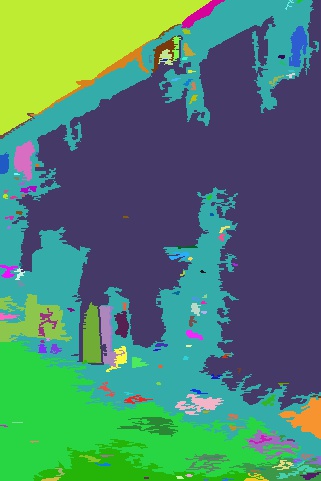}}
\end{minipage}
\begin{minipage}{0.072\linewidth}
\centerline{\includegraphics[scale=0.11322]{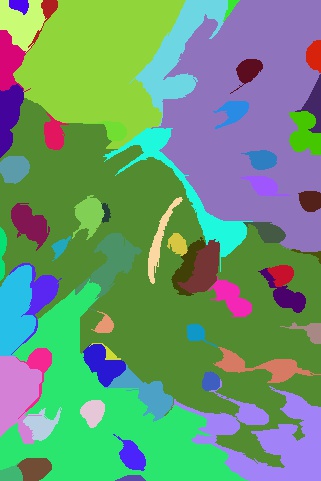}}
\end{minipage}
\\

\vspace{0.1cm}
\begin{minipage}{0.02\linewidth}
\centerline{(d)}
\end{minipage}
\begin{minipage}{0.16\linewidth}
\centerline{\includegraphics[scale=0.17]{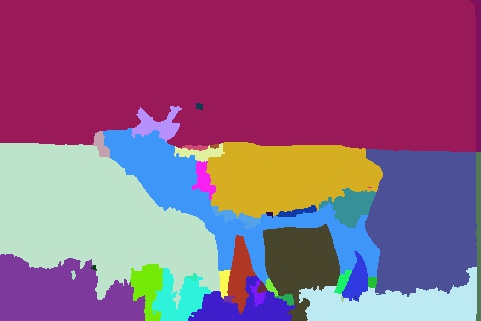}}
\end{minipage}
\begin{minipage}{0.16\linewidth}
\centerline{\includegraphics[scale=0.17]{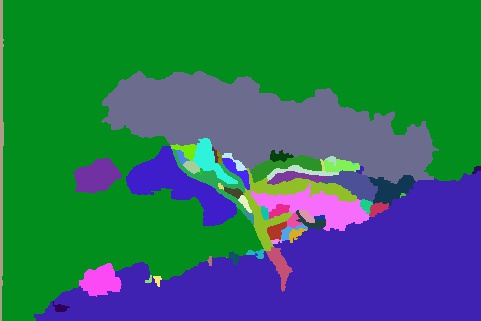}}
\end{minipage}
\begin{minipage}{0.16\linewidth}
\centerline{\includegraphics[scale=0.17]{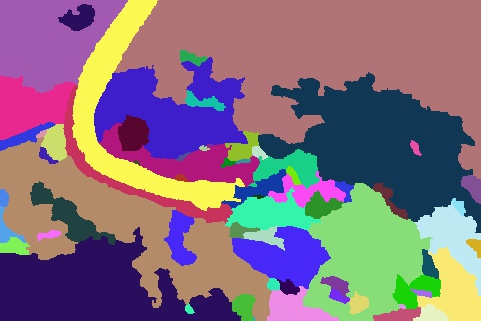}}
\end{minipage}
\begin{minipage}{0.16\linewidth}
\centerline{\includegraphics[scale=0.17]{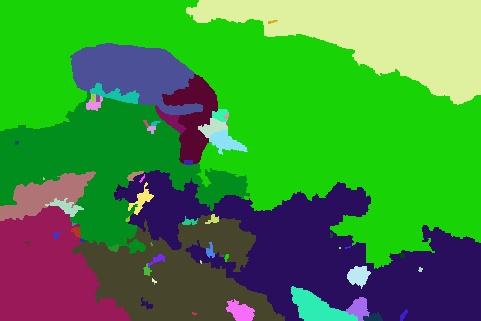}}
\end{minipage}
\begin{minipage}{0.072\linewidth}
\centerline{\includegraphics[scale=0.11322]{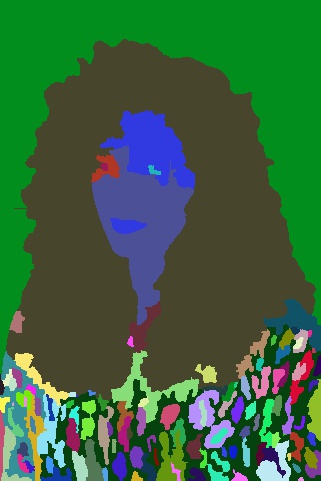}}
\end{minipage}
\begin{minipage}{0.072\linewidth}
\centerline{\includegraphics[scale=0.11322]{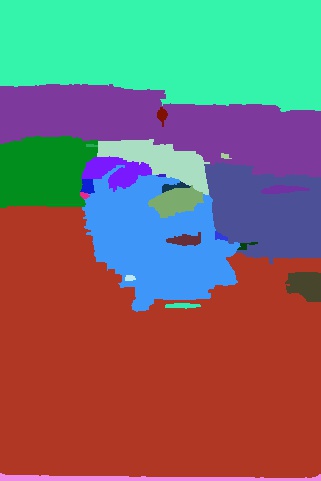}}
\end{minipage}
\begin{minipage}{0.072\linewidth}
\centerline{\includegraphics[scale=0.11322]{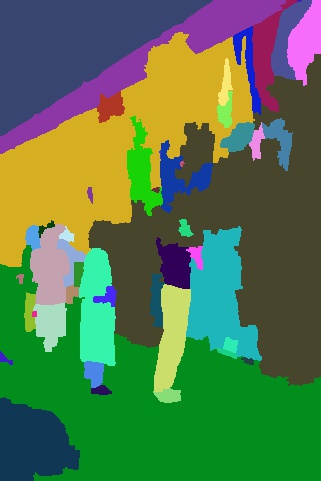}}
\end{minipage}
\begin{minipage}{0.072\linewidth}
\centerline{\includegraphics[scale=0.11322]{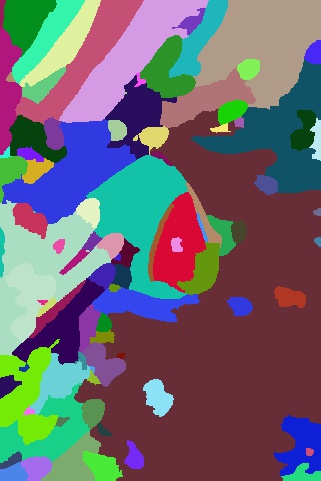}}
\end{minipage}
\\

\vspace{0.1cm}
\begin{minipage}{0.02\linewidth}
\centerline{(e)}
\end{minipage}
\begin{minipage}{0.16\linewidth}
\centerline{\includegraphics[scale=0.17]{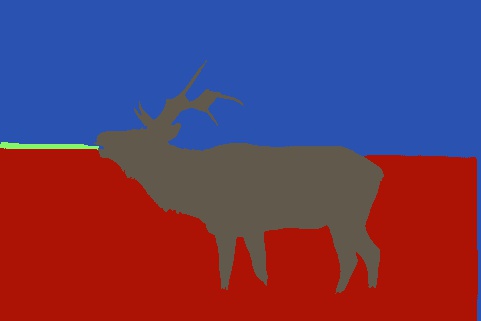}}
\end{minipage}
\begin{minipage}{0.16\linewidth}
\centerline{\includegraphics[scale=0.17]{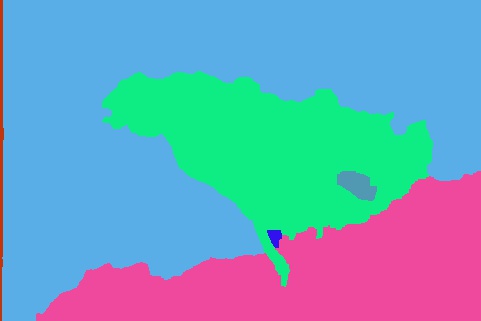}}
\end{minipage}
\begin{minipage}{0.16\linewidth}
\centerline{\includegraphics[scale=0.17]{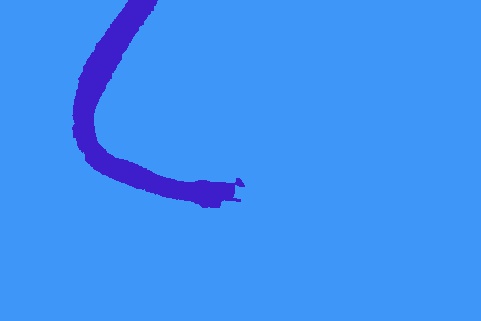}}
\end{minipage}
\begin{minipage}{0.16\linewidth}
\centerline{\includegraphics[scale=0.17]{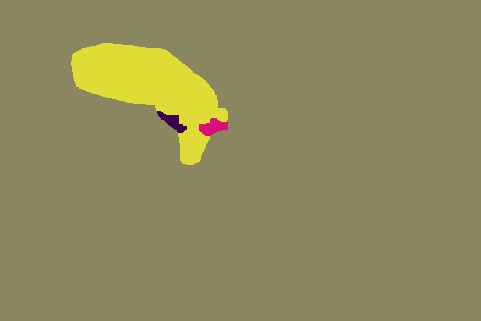}}
\end{minipage}
\begin{minipage}{0.072\linewidth}
\centerline{\includegraphics[scale=0.11322]{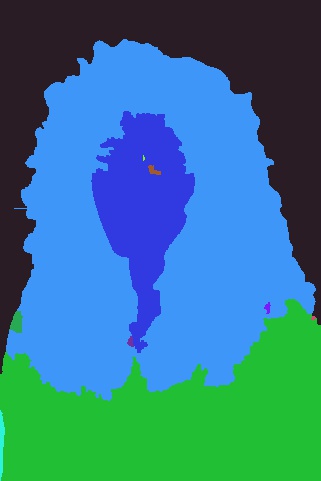}}
\end{minipage}
\begin{minipage}{0.072\linewidth}
\centerline{\includegraphics[scale=0.11322]{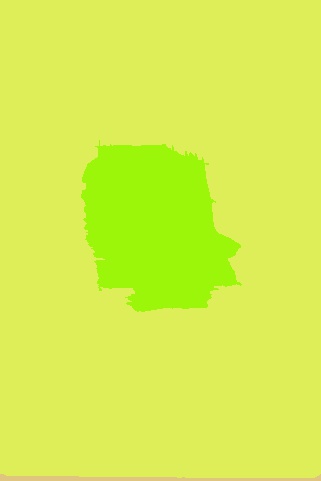}}
\end{minipage}
\begin{minipage}{0.072\linewidth}
\centerline{\includegraphics[scale=0.11322]{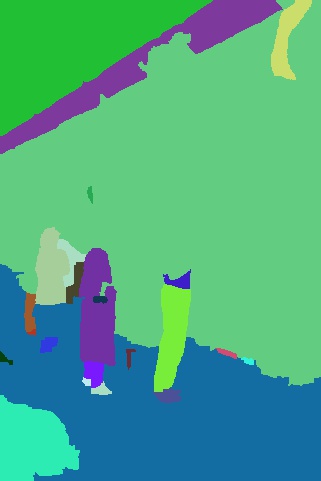}}
\end{minipage}
\begin{minipage}{0.072\linewidth}
\centerline{\includegraphics[scale=0.11322]{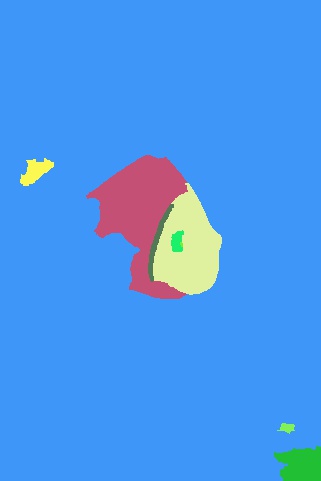}}
\end{minipage}
\caption{Qualitative results on the BSDS500 dataset~\citep{martin2001database}. (a) Image; (b) Ground truth; (c) FH \citep{ graph2004}; (d) DIC \citep{dic2020}; (e) Proposed method.}
\label{fig:result_bsds500}
\end{center}
\end{figure*}

\begin{figure*}[!t] 
\begin{center}
\begin{minipage}{0.015\linewidth}
\centerline{(a)}
\end{minipage}
\begin{minipage}{0.172\linewidth}
\centerline{\includegraphics[scale=0.1672]{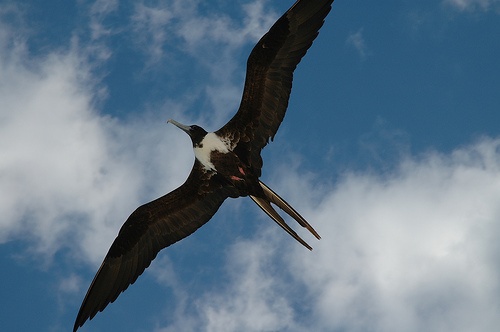}}
\end{minipage}
\begin{minipage}{0.111\linewidth}
\centerline{\includegraphics[scale=0.11115]{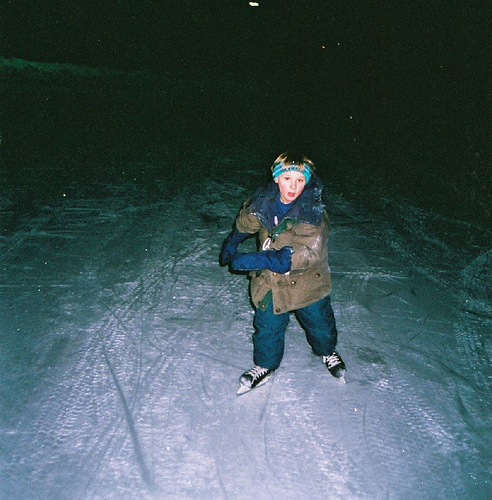}}
\end{minipage}
\begin{minipage}{0.151\linewidth}
\centerline{\includegraphics[scale=0.1482]{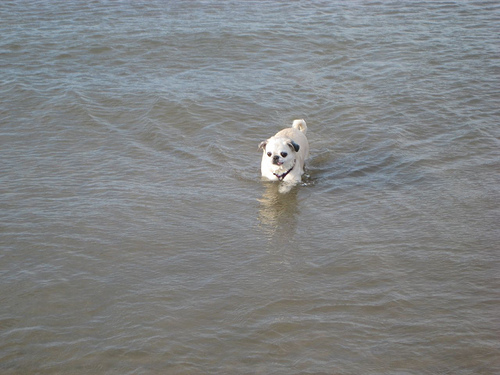}}
\end{minipage}
\begin{minipage}{0.167\linewidth}
\centerline{\includegraphics[scale=0.2242]{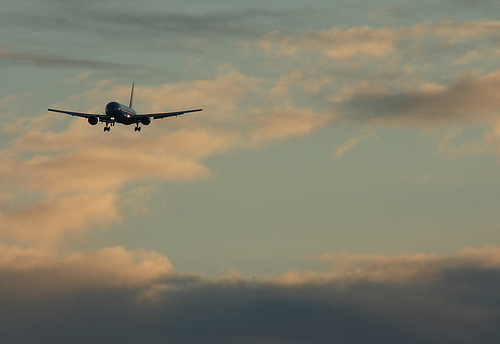}}
\end{minipage}
\begin{minipage}{0.15\linewidth}
\centerline{\includegraphics[scale=0.1482]{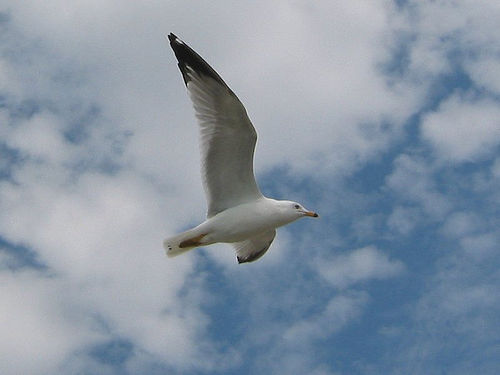}}
\end{minipage}
\begin{minipage}{0.172\linewidth}
\centerline{\includegraphics[scale=0.9101]{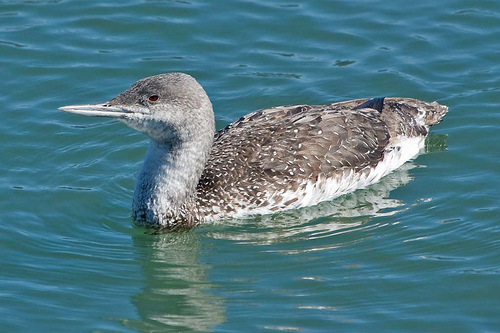}}
\end{minipage}
\\

\vspace{0.1cm}
\begin{minipage}{0.015\linewidth}
\centerline{(b)}
\end{minipage}
\begin{minipage}{0.172\linewidth}
\centerline{\includegraphics[scale=0.1672]{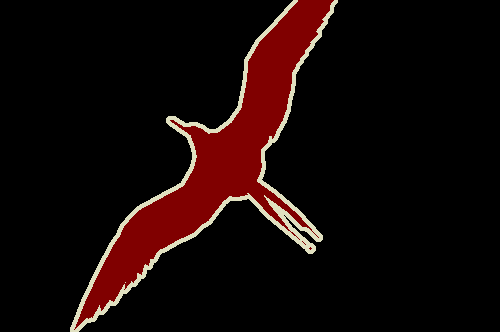}}
\end{minipage}
\begin{minipage}{0.111\linewidth}
\centerline{\includegraphics[scale=0.11115]{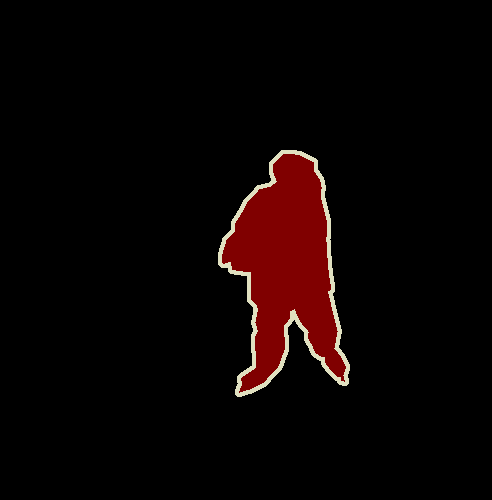}}
\end{minipage}
\begin{minipage}{0.151\linewidth}
\centerline{\includegraphics[scale=0.1482]{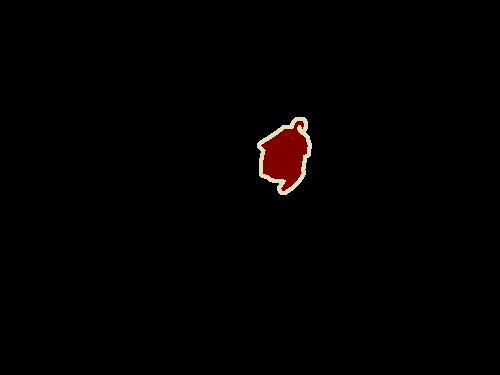}}
\end{minipage}
\begin{minipage}{0.167\linewidth}
\centerline{\includegraphics[scale=0.16055]{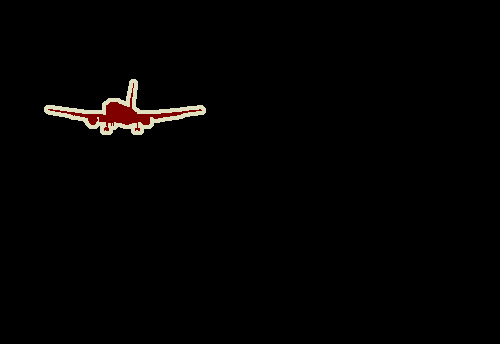}}
\end{minipage}
\begin{minipage}{0.15\linewidth}
\centerline{\includegraphics[scale=0.1482]{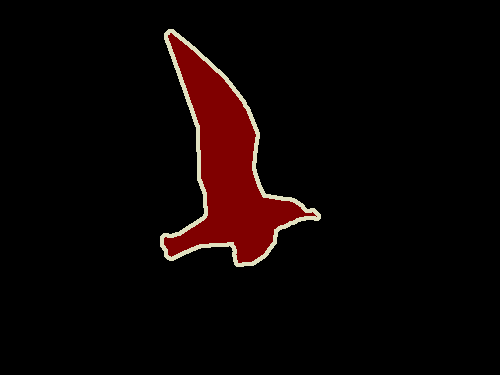}}
\end{minipage}
\begin{minipage}{0.172\linewidth}
\centerline{\includegraphics[scale=0.16625]{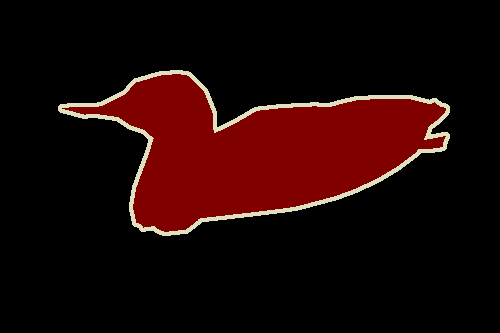}}
\end{minipage}
\\

\vspace{0.1cm}
\begin{minipage}{0.015\linewidth}
\centerline{(c)}
\end{minipage}
\begin{minipage}{0.172\linewidth}
\centerline{\includegraphics[scale=0.1672]{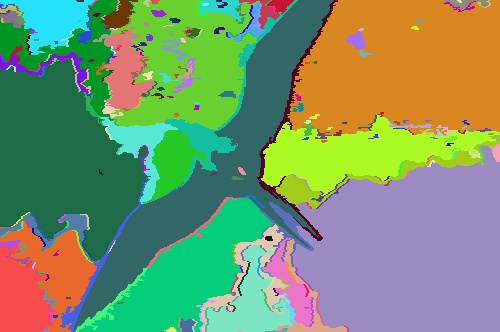}}
\end{minipage}
\begin{minipage}{0.111\linewidth}
\centerline{\includegraphics[scale=0.11115]{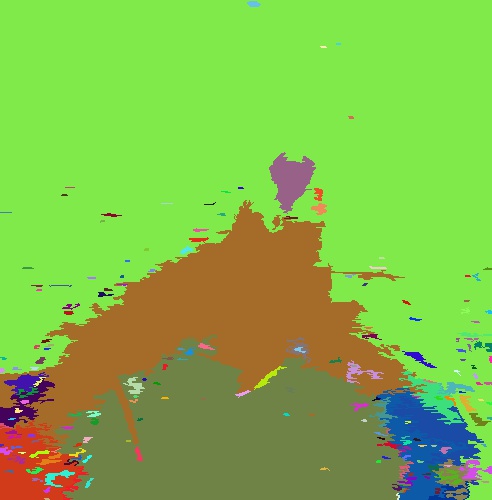}}
\end{minipage}
\begin{minipage}{0.151\linewidth}
\centerline{\includegraphics[scale=0.1482]{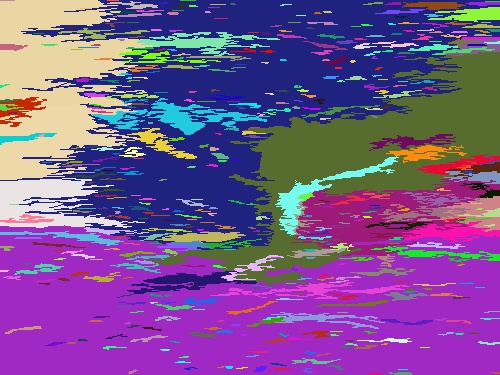}}
\end{minipage}
\begin{minipage}{0.167\linewidth}
\centerline{\includegraphics[scale=0.16055]{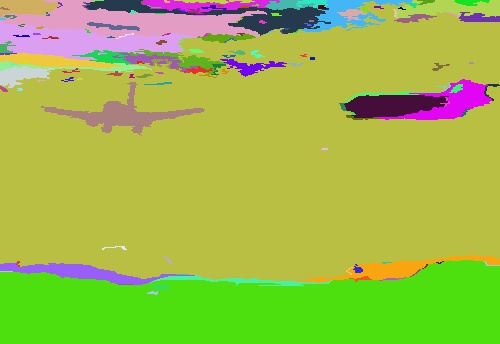}}
\end{minipage}
\begin{minipage}{0.15\linewidth}
\centerline{\includegraphics[scale=0.1482]{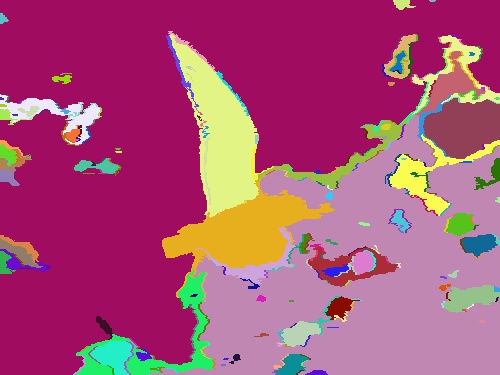}}
\end{minipage}
\begin{minipage}{0.172\linewidth}
\centerline{\includegraphics[scale=0.16625]{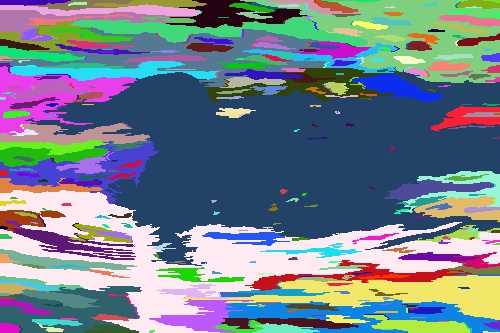}}
\end{minipage}
\\

\vspace{0.1cm}
\begin{minipage}{0.015\linewidth}
\centerline{(d)}
\end{minipage}
\begin{minipage}{0.172\linewidth}
\centerline{\includegraphics[scale=0.1672]{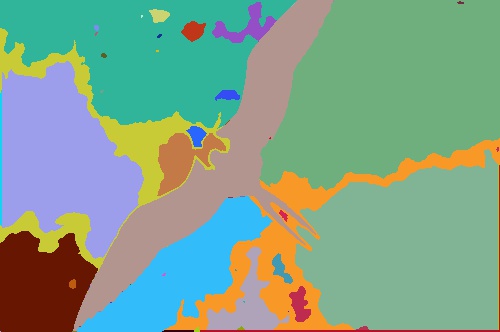}}
\end{minipage}
\begin{minipage}{0.111\linewidth}
\centerline{\includegraphics[scale=0.11115]{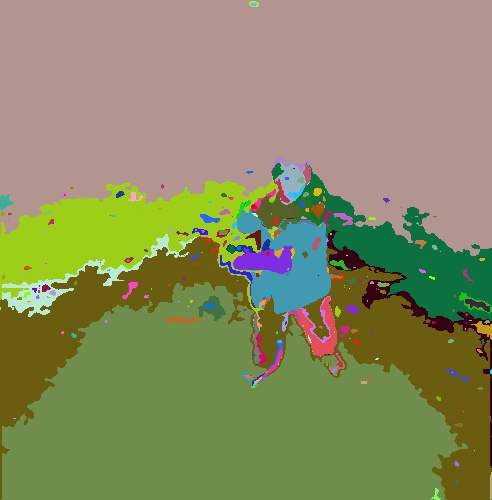}}
\end{minipage}
\begin{minipage}{0.151\linewidth}
\centerline{\includegraphics[scale=0.1482]{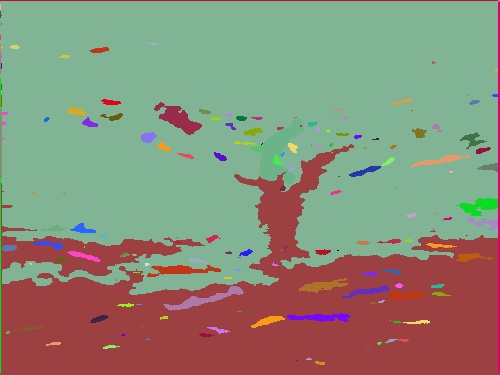}}
\end{minipage}
\begin{minipage}{0.167\linewidth}
\centerline{\includegraphics[scale=0.16055]{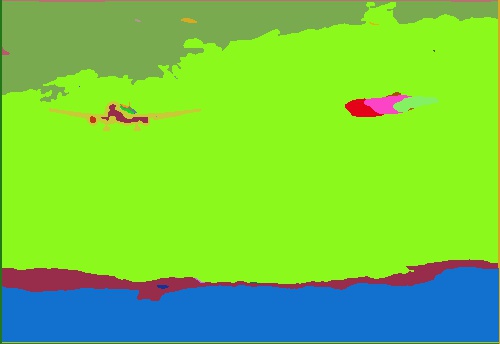}}
\end{minipage}
\begin{minipage}{0.15\linewidth}
\centerline{\includegraphics[scale=0.1482]{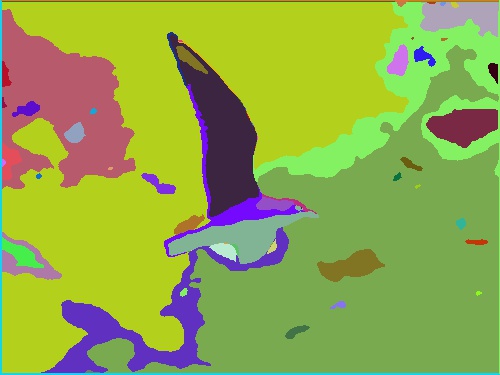}}
\end{minipage}
\begin{minipage}{0.172\linewidth}
\centerline{\includegraphics[scale=0.16625]{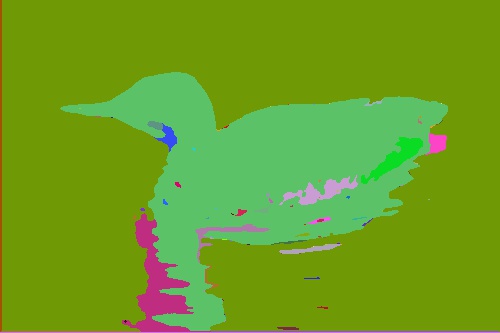}}
\end{minipage}
\\

\vspace{0.1cm}
\begin{minipage}{0.015\linewidth}
\centerline{(e)}
\end{minipage}
\begin{minipage}{0.172\linewidth}
\centerline{\includegraphics[scale=0.1672]{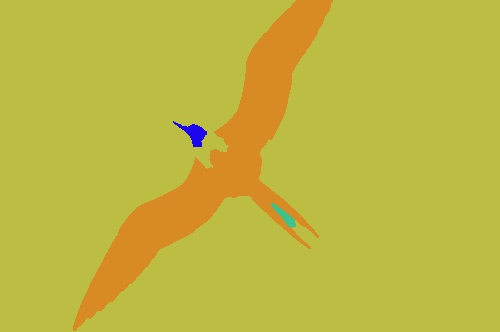}}
\end{minipage}
\begin{minipage}{0.111\linewidth}
\centerline{\includegraphics[scale=0.11115]{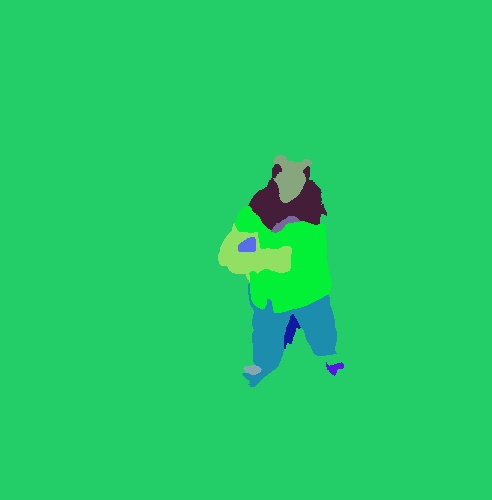}}
\end{minipage}
\begin{minipage}{0.151\linewidth}
\centerline{\includegraphics[scale=0.1482]{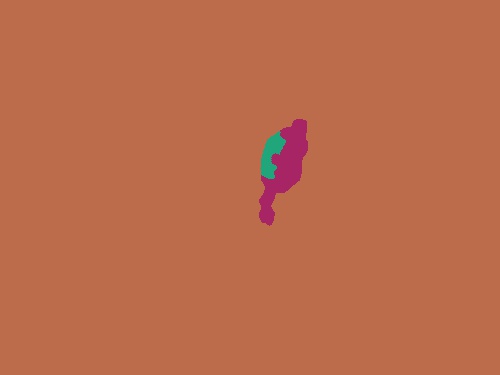}}
\end{minipage}
\begin{minipage}{0.167\linewidth}
\centerline{\includegraphics[scale=0.16055]{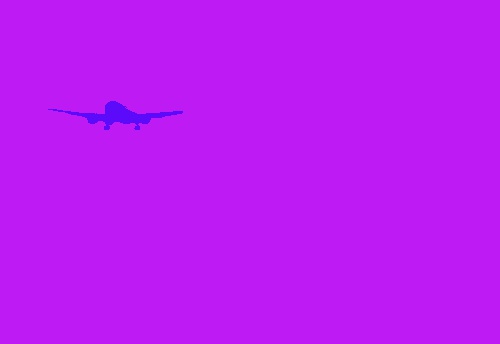}}
\end{minipage}
\begin{minipage}{0.15\linewidth}
\centerline{\includegraphics[scale=0.1482]{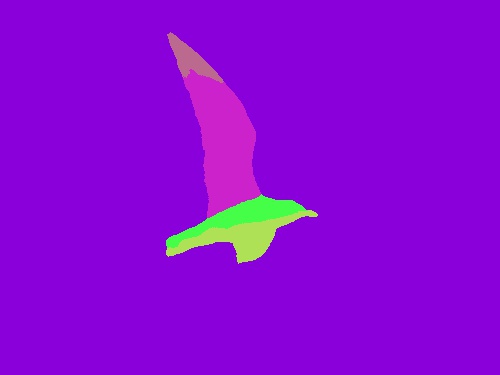}}
\end{minipage}
\begin{minipage}{0.172\linewidth}
\centerline{\includegraphics[scale=0.16625]{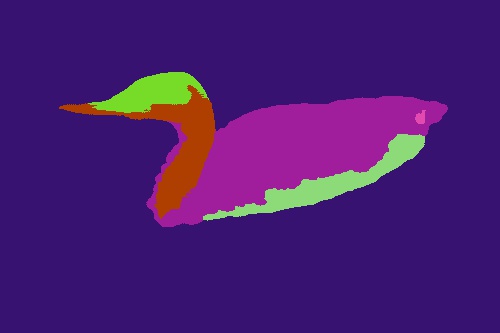}}
\end{minipage}
\caption{Qualitative results on the PASCAL VOC 2012 dataset~\citep{Everingham2010}. (a) Image; (b) Ground truth; (c) FH~\citep{graph2004}; (d) DFC~\citep{kim2020unsupervised}; (e) Proposed method.}
\label{fig:result_pascal}
\end{center}
\end{figure*}

\begin{table*}[!t]
\small
\begin{center}
\begin{minipage}{0.9\linewidth}
\caption{Ablation study on components of the proposed framework using the BSDS300 dataset~\citep{martin2001database}.}
\label{tab:ablation}
\begin{tabu}{X[c,m] X[c,m] X[c,m] X[c,m] c | X[c,m] X[c,m] X[c,m] X[c,m]} 
 \hline
 Baseline & ECA & $\calL_{global}$ & $\calL_{rec}$ & Post-processing & PRI & VoI & GCE & BDE \\ 
 \hline \hline
 $\surd$ & & & & & 0.801 & 1.931 & 0.152 & 11.162 \\ 
 $\surd$ & $\surd$ & & & & 0.813 & 1.792 & 0.160  & 10.914 \\
 $\surd$ & $\surd$ & $\surd$ & & & 0.821 & 1.786 & 0.162 & 10.764 \\
 $\surd$ & $\surd$ & $\surd$ & $\surd$ & & 0.822 & 1.755 & 0.162 & 10.694 \\ 
 $\surd$ & $\surd$ & $\surd$ & $\surd$ & $\surd$ & 0.830 & 1.613 & 0.170 & 10.256 \\ 
 \hline
\end{tabu}
\end{minipage}
\end{center}
\end{table*}

\subsection{Result}
For the BSDS dataset~\citep{martin2001database}, we utilize five metrics that are Segmentation Covering (SC), Probabilistic Rand Index (PRI), Variation of Information (VoI), Global Consistency Error (GCE), and Boundary Displacement Error (BDE) to compare results quantitatively. Considering SC and PRI, higher scores represent better results. For VoI, GCE, and BDE, lower values denote better segmentation. Following previous works~\citep{xia2017w}, we use PRI, VoI, GCE, and BDE for the BSDS300 dataset and SC, PRI, and VoI for the BSDS500 dataset. As the BSDS300 dataset is a part of the BSDS500 dataset, corresponding results are quite related. 

\tref{tab:result_300} shows quantitative results on the BSDS300 dataset~\citep{martin2001database}. We compare the performance of the proposed method to those of previous methods~\citep{kim2012learning, li2012segmentation, xia2017w, fu2015robust, dic2020}. In the table, we use boldface and underlines to denote the best and the second-best scores, respectively. The proposed method achieves the best scores in PRI, VoI, and BDE and the third-best score in GCE. The DIC method by~\cite{dic2020} achieves the second-best scores in PRI, GCE, and BDE. The CCP algorithm achieves the best score in GCE~\citep{fu2015robust}. The gPb-owt-ucm method by~\cite{arbelaez2009contours} achieves the second-best score in VoI.

Quantitative results on the BSDS500 dataset are shown in~\tref{tab:result_500}. The proposed method is compared to previous works~\citep{wang2017unsupervised, Kanezaki2018, Lin2020, xia2017w, dic2020}. The proposed method achieves the best scores in all metrics (SC, PRI, and VoI). The DIC method by~\cite{dic2020} achieves the second-best scores in SC and PRI. The gPb-owt-ucm method by~\cite{arbelaez2009contours} achieves the second-best score in VoI.

\begin{figure}[!t] \begin{center}
\begin{minipage}{0.3\linewidth}
\centerline{\includegraphics[scale=0.22]{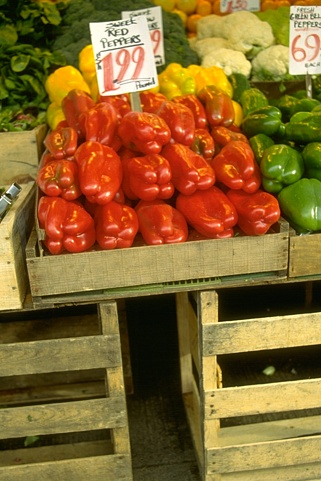}}
\end{minipage}
\begin{minipage}{0.3\linewidth}
\centerline{\includegraphics[scale=0.22]{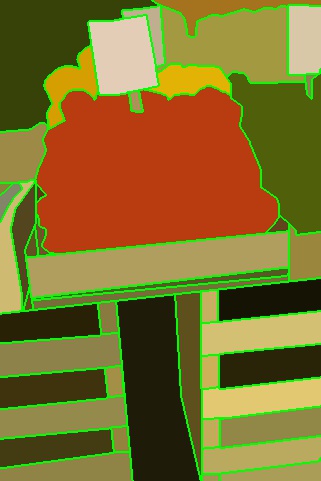}}
\end{minipage}
\begin{minipage}{0.3\linewidth}
\centerline{\includegraphics[scale=0.22]{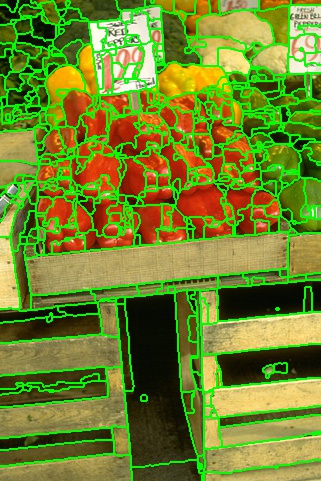}}
\end{minipage}
\\

\vspace{0.1cm}
\begin{minipage}{0.3\linewidth}
\centerline{{\footnotesize Image}}
\end{minipage}
\begin{minipage}{0.3\linewidth}
\centerline{{\footnotesize Ground truth}}
\end{minipage}
\begin{minipage}{0.3\linewidth}
\centerline{{\footnotesize Superpixels}}
\end{minipage}
\\

\vspace{0.1cm}
\begin{minipage}{0.3\linewidth}
\centerline{\includegraphics[scale=0.22]{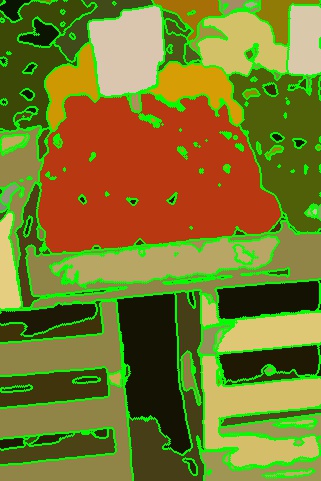}}
\end{minipage}
\begin{minipage}{0.3\linewidth}
\centerline{\includegraphics[scale=0.22]{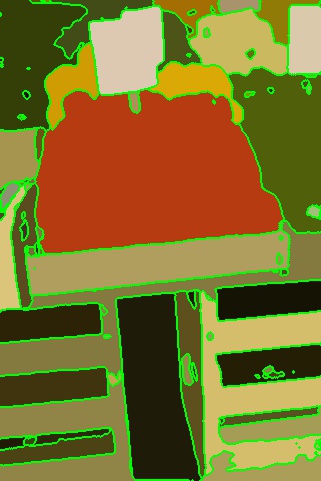}}
\end{minipage}
\begin{minipage}{0.3\linewidth}
\centerline{\includegraphics[scale=0.22]{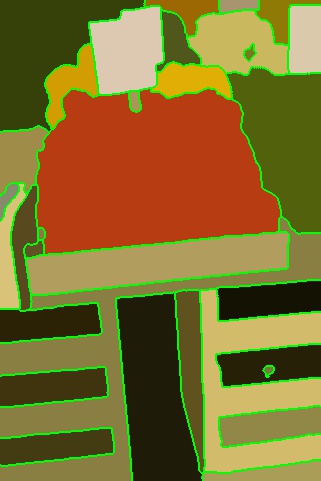}}
\end{minipage}
\\

\vspace{0.1cm}
\begin{minipage}{0.3\linewidth}
\centerline{{\footnotesize $T=50$}}
\end{minipage}
\begin{minipage}{0.3\linewidth}
\centerline{{\footnotesize $T=100$}}
\end{minipage}
\begin{minipage}{0.3\linewidth}
\centerline{{\footnotesize $T=150$}}
\end{minipage}
\caption{Qualitative results at the varying number of iterations.}
\label{fig:ablation_result}
\end{center} \end{figure}

\begin{figure}[!t] \begin{center}
\begin{minipage}{0.45\linewidth}
\centerline{\includegraphics[scale=0.21]{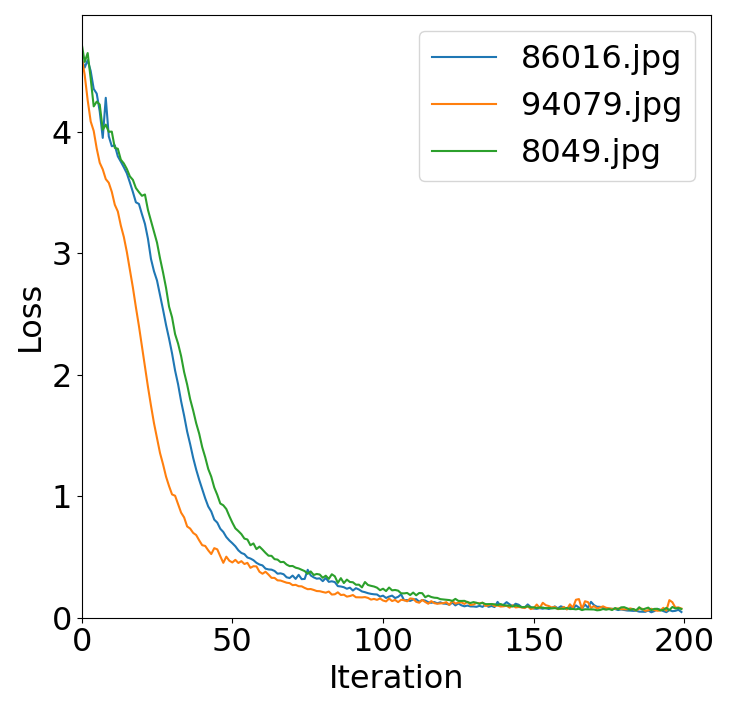}}
\end{minipage}
\begin{minipage}{0.45\linewidth}
\centerline{\includegraphics[scale=0.21]{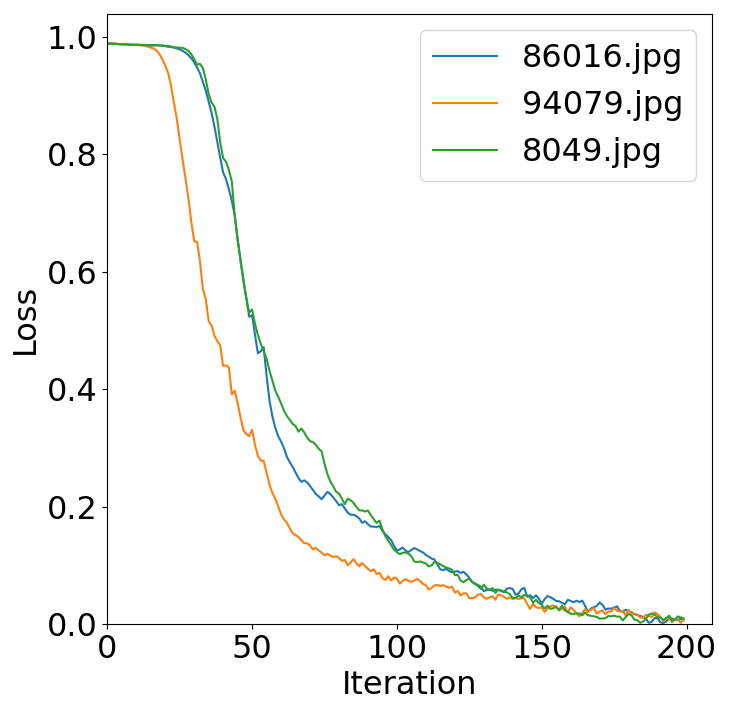}}
\end{minipage}
\\

\vspace{0.1cm}
\begin{minipage}{0.45\linewidth}
\centerline{{\footnotesize (a) $\calL_{local}$}}
\end{minipage}
\begin{minipage}{0.45\linewidth}
\centerline{{\footnotesize (b) $\calL_{global}$}}
\end{minipage}
\\

\vspace{0.1cm}
\begin{minipage}{0.45\linewidth}
\centerline{\includegraphics[scale=0.21]{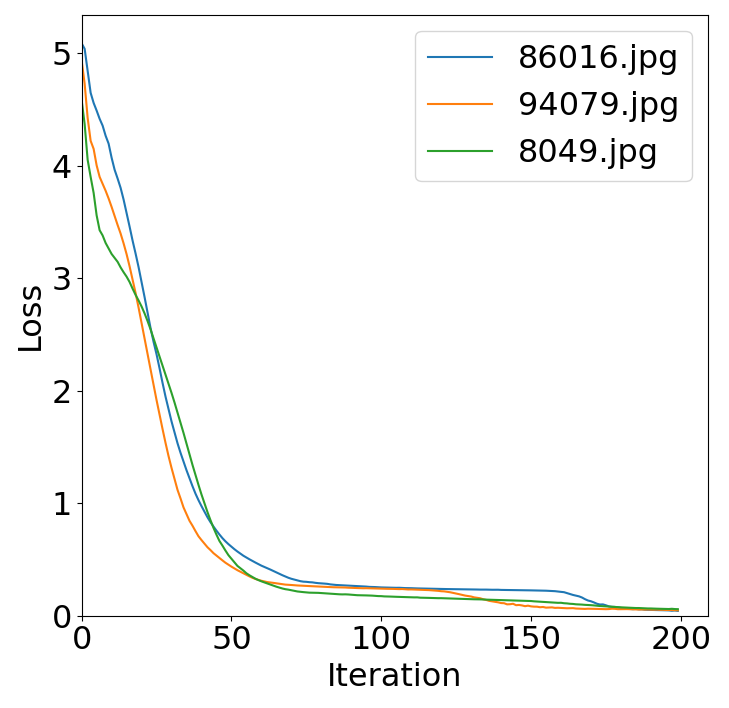}}
\end{minipage}
\\

\vspace{0.1cm}
\begin{minipage}{0.45\linewidth}
\centerline{{\footnotesize (c) $\calL_{rec}$}}
\end{minipage}
\caption{Change of loss values during training.}
\label{fig:ablation_loss}
\end{center} \end{figure}

We evaluated the proposed method on the PASCAL VOC 2012 dataset~\citep{Everingham2010} and computed the mean Intersection over Union (mIoU) for quantitative comparison. The results are shown in~\tref{tab:result_pascal}, which demonstrate that the proposed method outperforms all other methods~\citep{graph2004, IIC2019, Kanezaki2018, kim2020unsupervised}.

\fref{fig:result_bsds500} shows qualitative results using the BSDS500 dataset. Each row, from top to bottom, shows an input image, ground truth, and the results of FH~\citep{graph2004}, DIC~\citep{dic2020}, and the proposed method. \fref{fig:result_pascal} shows qualitative results using the PASCAL VOC 2012 dataset. Each row, from top to bottom, shows an input image, ground truth, and the results of FH~\citep{graph2004}, DFC~\citep{kim2020unsupervised}, and the proposed method. Both qualitative results demonstrate that the proposed method achieves accurate segmentation compared to others~\citep{graph2004, dic2020, kim2020unsupervised}. Moreover, the results show that the proposed method segments images into a reasonable number of clusters while other methods often produce unnecessarily over-segmented results.

\tref{tab:ablation} presents an ablation study on the components of the proposed framework using the BSDS300 dataset. The baseline model in~\tref{tab:ablation} represents the framework without the attention mechanism (ECA), image reconstruction modules, and post-processing step. Also, the baseline model is trained using only $\calL_{local}$. The first and second rows show the results of the baseline model and the model with the attention module (ECA), respectively. The third and fourth rows show the results of including $\calL_{global}$ and $\calL_{rec}$. The last row shows the result of the proposed framework. Please note that the scores are different from those in~\tref{tab:result_300} because of OIS. The results in~\tref{tab:result_300} include OIS, whereas those in~\tref{tab:ablation} do not. In this ablation study, the number of superpixels $K$ is fixed at 100 for all the images.

To analyze the optimization process, we show the results at varying numbers of iterations in~\fref{fig:ablation_result}. The figure shows the input image, ground truth, superpixel segmentation result, and results of the proposed network at 50, 100, and 150 iterations.  The results in this figure do not include post-processing. We also demonstrate the convergence of loss terms by examining three different images. \fref{fig:ablation_loss} shows the change in loss values during training. 

\subsection{Extension to Unsupervised Semantic Segmentation}
While the proposed method is designed for unsupervised image segmentation, we show the additional value of the proposed method by applying it to unsupervised semantic segmentation. Specifically, we fuse the proposed method with one of the state-of-the-art methods, STEGO~\citep{hamilton2022unsupervised}, in unsupervised semantic segmentation. We then demonstrate that the fused method outperforms the previous state-of-the-art performance using the COCO-Stuff dataset~\citep{caesar2018coco}.

The extended method first applies the proposed method to each image to segment the image into multiple regions. Then, the image is cropped into multiple patches using the bounding boxes of segmented regions. The features of the cropped images are obtained by forward-propagating them through the pre-trained PVTv2-B5 backbone~\citep{wang2022pvt} and by processing mask-based pooling. The backbone is pre-trained using the self-supervised learning method by~\cite{caron2021emerging} without any human annotations. The mask-based pooling aggregates extracted features over each segmented region. The aggregated feature is then processed by a learnable segmentation head to reduce the dimension. Finally, the centers of the $k$ clusters are determined using the outputs of the segmentation head, where $k$ is the number of semantic categories in the dataset. For clustering, the similarities are computed by the cosine distance. Unlike STEGO~\citep{hamilton2022unsupervised}, conditional random field (CRF)-based refinement is not employed since the proposed method produces high-quality and detailed segmentation masks.

For evaluation and visualization, the Hungarian matching algorithm is applied to match clusters to ground-truth labels. Following STEGO~\citep{hamilton2022unsupervised}, the pre-trained backbone is frozen during training while the segmentation head is trained using the contrastive loss function. Positive and negative samples are obtained by finding $k$-nearest neighbors and by random sampling, respectively. We refer readers to STEGO~\citep{hamilton2022unsupervised} for further details.
  
\begin{table}[!t]
\small
\begin{center}
\begin{minipage}{0.98\linewidth}
\caption{Quantitative results on the COCO-Stuff validation dataset~\citep{caesar2018coco}.}
\label{tab:result_cocostuff}
\renewcommand{\arraystretch}{1.1} 
\begin{tabular}{>{\centering}m{0.58\textwidth}|>{\centering}m{0.15\textwidth}|>{\centering\arraybackslash}m{0.15\textwidth}} 
\hline
Methods & Accuracy & mIoU   \\ 
\hline\hline
ResNet50~\citep{he2016deep}  & 24.6 & 8.9   \\ 
MoCoV2~\citep{chen2020improved}& 25.2 & 10.4   \\ 
DINO~\citep{caron2021emerging} & 30.5 & 9.6 \\ 
Deep Cluster~\citep{caron2018deep} & 19.9 & -   \\ 
SIFT~\citep{lowe1999object}& 20.2 & -   \\ 
AC~\citep{ouali2020autoregressive} & 30.8 & -   \\ 
InMARS~\citep{mirsadeghi2021unsupervised} & 31.0 & -   \\ 
IIC~\citep{ji2019invariant}& 21.8 & 6.7   \\ 
MDC~\citep{cho2021picie} & 32.2 & 9.8  \\ 
PiCIE~\citep{cho2021picie} & 41.1 & 13.8   \\ 
PiCIE + H~\citep{cho2021picie} & 50.0 & 14.4    \\  
STEGO~\citep{hamilton2022unsupervised} & \underline{56.9} & \underline{28.2}    \\
\hline
Proposed & \textbf{59.1}  & \textbf{33.6}    \\
\hline
\end{tabular}
\end{minipage}
\end{center}
\end{table}

\begin{figure*}[!t] 
\begin{center}
\begin{minipage}{0.02\linewidth}
\centerline{(a)}
\end{minipage}
\begin{minipage}{0.13\linewidth}
\centerline{\includegraphics[scale=0.2087]{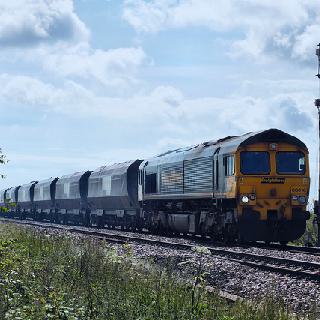}}
\end{minipage}
\begin{minipage}{0.13\linewidth}
\centerline{\includegraphics[scale=0.29]{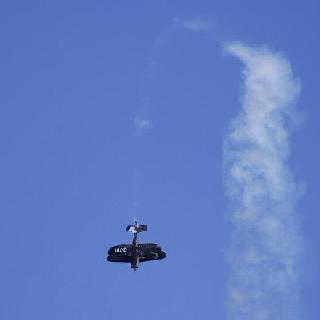}}
\end{minipage}
\begin{minipage}{0.13\linewidth}
\centerline{\includegraphics[scale=0.29]{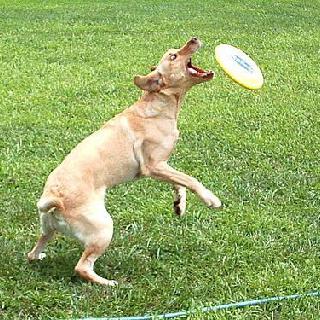}}
\end{minipage}
\begin{minipage}{0.13\linewidth}
\centerline{\includegraphics[scale=0.29]{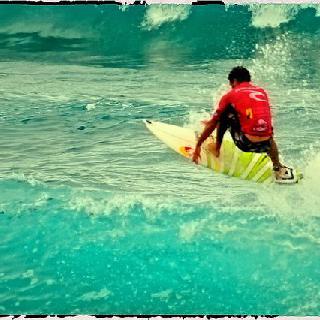}} 
\end{minipage}
\begin{minipage}{0.13\linewidth}
\centerline{\includegraphics[scale=0.29]{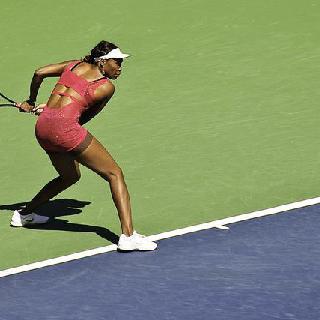}}
\end{minipage}
\begin{minipage}{0.13\linewidth}
\centerline{\includegraphics[scale=0.209]{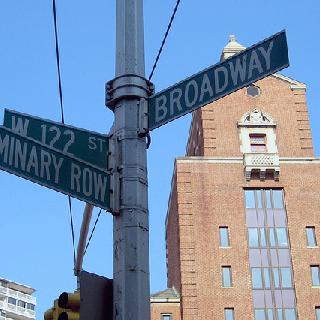}}
\end{minipage}
\begin{minipage}{0.13\linewidth}
\centerline{\includegraphics[scale=0.209]{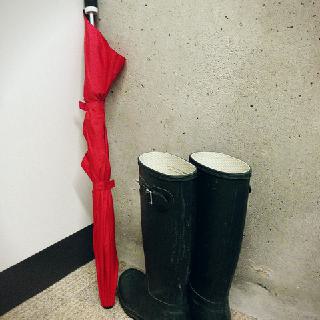}}
\end{minipage}
\\
\vspace{0.1cm}
\begin{minipage}{0.02\linewidth}
\centerline{(b)}
\end{minipage}
\begin{minipage}{0.13\linewidth}
\centerline{\includegraphics[scale=0.21]{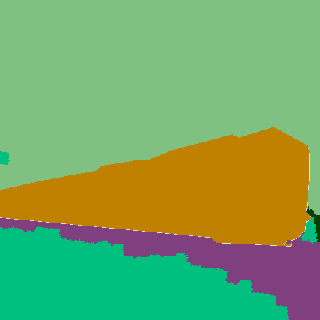}}
\end{minipage}
\begin{minipage}{0.13\linewidth}
\centerline{\includegraphics[scale=0.21]{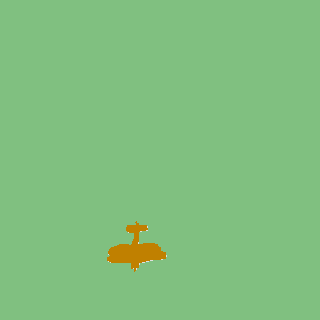}}
\end{minipage}
\begin{minipage}{0.13\linewidth}
\centerline{\includegraphics[scale=0.21]{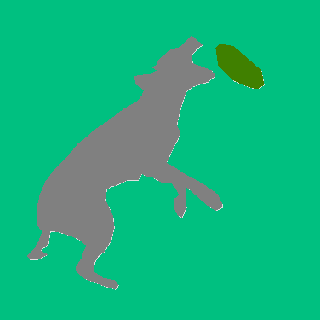}}
\end{minipage}
\begin{minipage}{0.13\linewidth}
\centerline{\includegraphics[scale=0.21]{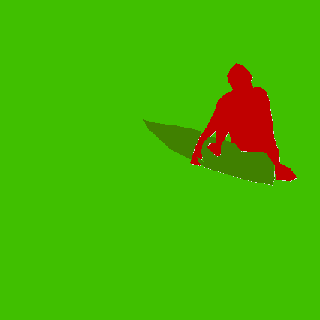}}
\end{minipage}
\begin{minipage}{0.13\linewidth}
\centerline{\includegraphics[scale=0.21]{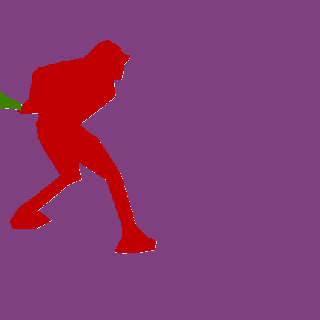}}
\end{minipage}
\begin{minipage}{0.13\linewidth}
\centerline{\includegraphics[scale=0.21]{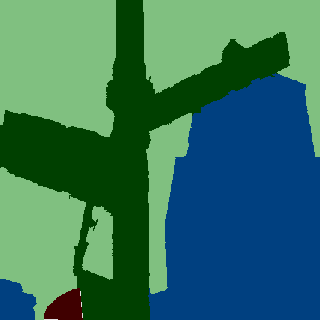}}
\end{minipage}
\begin{minipage}{0.13\linewidth}
\centerline{\includegraphics[scale=0.21]{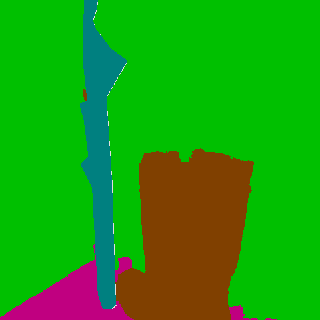}}
\end{minipage}
\\
\vspace{0.1cm}
\begin{minipage}{0.02\linewidth}
\centerline{(c)}
\end{minipage}
\begin{minipage}{0.13\linewidth}
\centerline{\includegraphics[scale=0.21]{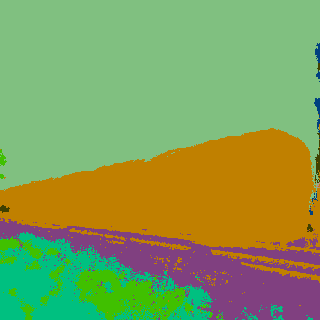}}
\end{minipage}
\begin{minipage}{0.13\linewidth}
\centerline{\includegraphics[scale=0.21]{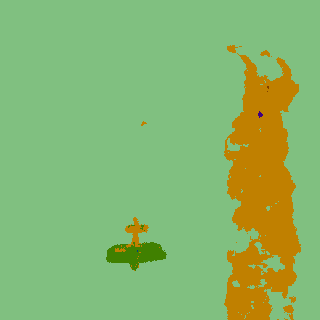}}
\end{minipage}
\begin{minipage}{0.13\linewidth}
\centerline{\includegraphics[scale=0.21]{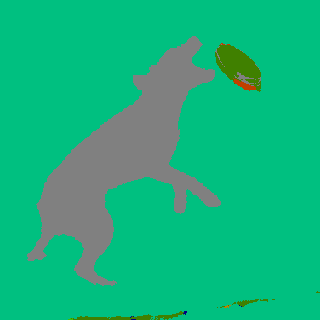}}
\end{minipage}
\begin{minipage}{0.13\linewidth}
\centerline{\includegraphics[scale=0.21]{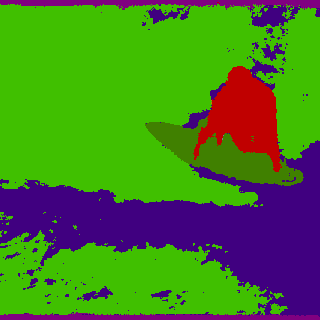}}
\end{minipage}
\begin{minipage}{0.13\linewidth}
\centerline{\includegraphics[scale=0.21]{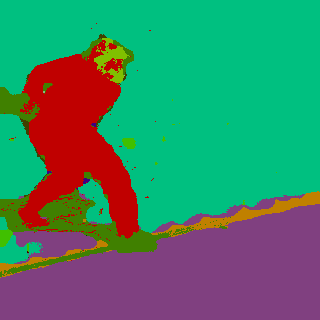}}
\end{minipage}
\begin{minipage}{0.13\linewidth}
\centerline{\includegraphics[scale=0.21]{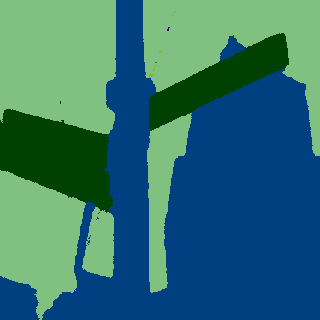}}
\end{minipage}
\begin{minipage}{0.13\linewidth}
\centerline{\includegraphics[scale=0.21]{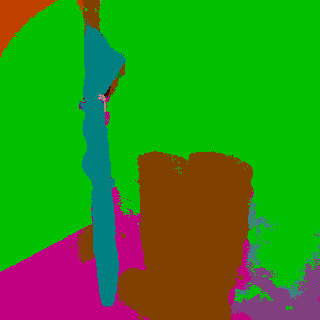}}
\end{minipage}
\\
\vspace{0.1cm}
\begin{minipage}{0.02\linewidth}
\centerline{(d)}
\end{minipage}
\begin{minipage}{0.13\linewidth}
\centerline{\includegraphics[scale=0.21]{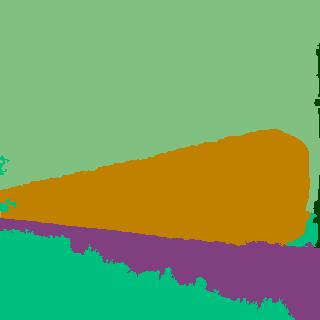}}
\end{minipage}
\begin{minipage}{0.13\linewidth}
\centerline{\includegraphics[scale=0.21]{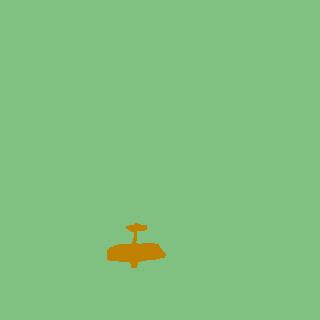}}
\end{minipage}
\begin{minipage}{0.13\linewidth}
\centerline{\includegraphics[scale=0.21]{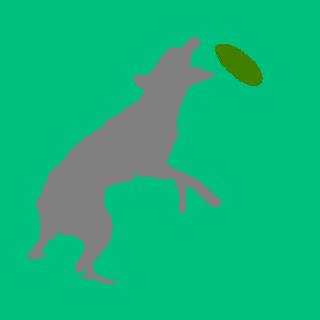}}
\end{minipage}
\begin{minipage}{0.13\linewidth}
\centerline{\includegraphics[scale=0.21]{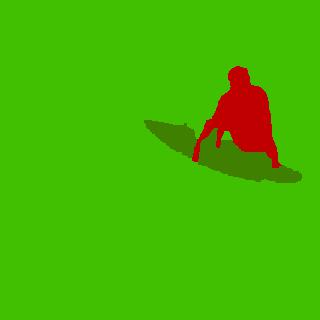}}
\end{minipage}
\begin{minipage}{0.13\linewidth}
\centerline{\includegraphics[scale=0.21]{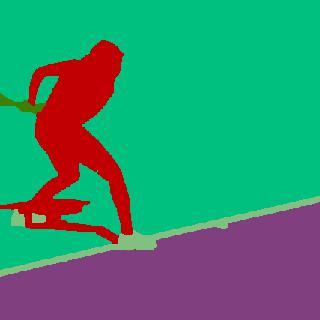}}
\end{minipage}
\begin{minipage}{0.13\linewidth}
\centerline{\includegraphics[scale=0.21]{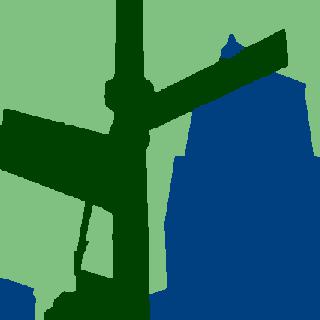}}
\end{minipage}
\begin{minipage}{0.13\linewidth}
\centerline{\includegraphics[scale=0.21]{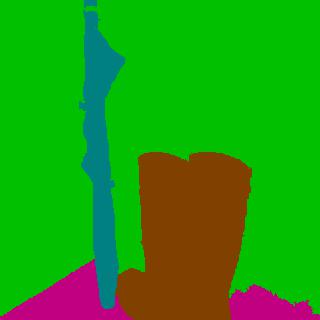}}
\end{minipage}
\\
\vspace{0.2cm}
\begin{minipage}{0.15\linewidth}
\centerline{\includegraphics[scale=0.34]{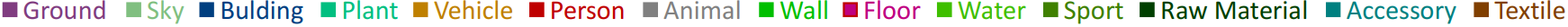}}
\end{minipage}
\caption{Qualitative results on the COCO-Stuff dataset~\citep{caesar2018coco}. (a) Image; (b) Ground truth; (c) STEGO~\citep{hamilton2022unsupervised}; (d) Proposed method.}
\label{fig:result_cocostuff}
\end{center}
\end{figure*}

Following previous literature~\citep{cho2021picie}, the fused method is evaluated using the 27 classes in the COCO-Stuff dataset~\citep{caesar2018coco}. Quantitative results demonstrate that the fused method outperforms the previous state-of-the-art method by achieving 59.1\% accuracy and 33.6 mIoU, as shown in~\tref{tab:result_cocostuff}. Qualitative results are shown in~\fref{fig:result_cocostuff}. The key difference between the fused method and STEGO~\citep{hamilton2022unsupervised} is that the fused method utilizes aggregated features over each segmented region while STEGO uses pixel-level features. We believe that given high-quality image segmentation results, the aggregated features are more consistent within each class and more discriminative between categories than pixel-level features. Moreover, since the fused method utilizes high-quality and detailed segmentation results from the proposed method, it preserves the boundaries of regions/objects better than the previous method~\citep{hamilton2022unsupervised}, as shown in~\fref{fig:result_cocostuff}.

\section{CONCLUSION}
We presented a novel pixel-level clustering framework for unsupervised image segmentation. The framework includes four feature embedding modules, a feature statistics computing component, two image reconstruction modules, and a superpixel segmentation algorithm. The proposed network is trained by ensuring consistency within each superpixel, utilizing feature similarity/dissimilarity between neighboring superpixels, and comparing an input image to reconstructed images from encoded features. Additionally, we included a post-processing method to overcome limitations caused by superpixels. Furthermore, we presented an extension of the proposed method for unsupervised semantic segmentation, which demonstrates the additional value of our approach. The experimental results indicate that the proposed method outperforms previous state-of-the-art methods. As the proposed framework can segment any given input image without any ground truth annotations or pre-training, it can be utilized in various real-world scenarios. For instance, it can help robots grasp unseen objects or discover new objects from a scene. Moreover, it can reduce the effort required for pixel-level annotation in supervised learning.

\section*{Acknowledgments}
This work was supported in part by the Korea Evaluation Institute of Industrial Technology (KEIT) Grant through the Korea Government (MOTIE) under Grant 20018635.

\bibliographystyle{elsarticle-harv}
\bibliography{mybibfile}

\end{document}